\documentclass[10pt,twocolumn,letterpaper]{article}

\usepackage[pagenumbers]{cvpr}
\def\benchEnglishOnly{}
\usepackage{iftex}
\ifPDFTeX
  \PackageError{preamble}{%
    This Chinese draft requires XeLaTeX.\MessageBreak
    Overleaf: Menu -> Compiler -> XeLaTeX}{}
\fi

\usepackage{fontspec}

\ifdefined\benchEnglishOnly\else
\usepackage{xeCJK}
\IfFontExistsTF{Songti SC}{%
  \setCJKmainfont[BoldFont={Heiti SC},ItalicFont={Songti SC}]{Songti SC}
  \setCJKsansfont{Heiti SC}
  \setCJKmonofont{Heiti SC}
}{%
  \IfFileExists{fonts/FandolSong-Regular.otf}{%
    \setCJKmainfont[Path=fonts/,BoldFont=FandolHei-Regular.otf,ItalicFont=FandolSong-Regular.otf]{FandolSong-Regular.otf}
    \setCJKsansfont[Path=fonts/]{FandolHei-Regular.otf}
    \setCJKmonofont[Path=fonts/]{FandolHei-Regular.otf}
  }{%
    \setCJKmainfont{FandolSong}[BoldFont=FandolHei]
    \setCJKsansfont{FandolHei}
    \setCJKmonofont{FandolHei}
  }
}

\fi

\usepackage{cuted}      % for 2-column strip teaser
\usepackage{currfile}   % for \currfilebase macro
\usepackage{caption}    % for \captionof
\usepackage{subcaption}
\usepackage{booktabs}
\usepackage{multirow}
\usepackage{tabularx}
\usepackage{threeparttable}
\usepackage{colortbl}
\usepackage{amsmath}
\usepackage{amssymb}
\usepackage{bm}
\usepackage{pifont}
\usepackage{xspace}

\definecolor{cgreen}{RGB}{34,139,34}
\definecolor{cred}{RGB}{197,48,48}
\definecolor{cgray}{RGB}{242,244,247}
\definecolor{cblue}{RGB}{43,108,176}

\newcommand{\cmark}{\textcolor{cgreen}{\ding{51}}}
\newcommand{\xmark}{\textcolor{cred}{\ding{55}}}

\newcommand{\bench}{BlenderFaceBench\xspace}
\newcommand{\fstrict}{\ensuremath{\text{F}@0.5\%}\xspace}
\newcommand{\floose}{\ensuremath{\text{F}@2\%}\xspace}
\newcommand{\gain}{\ensuremath{\mathrm{Gain}}\xspace}
\newcommand{\ident}{\ensuremath{\mathrm{Identity}}\xspace}
\newcommand{\initf}{\ensuremath{F_{\mathrm{init}}}\xspace}
\newcommand{\otherf}{\ensuremath{F_{\mathrm{other}}}\xspace}

\setlist[itemize]{nosep,leftmargin=1.3em,topsep=3pt}
\setlist[enumerate]{nosep,leftmargin=1.4em,topsep=3pt}
\usepackage{float}
\usepackage{array}
\usepackage{makecell}
\usepackage{tcolorbox}
\tcbuselibrary{skins}

\definecolor{cicolor}{RGB}{95,108,125}
\definecolor{slashcolor}{RGB}{140,154,172}
\definecolor{promptbg}{RGB}{250,251,253}
\definecolor{prompthead}{RGB}{238,243,249}
\definecolor{promptborder}{RGB}{188,202,218}
\definecolor{prompttitle}{RGB}{32,68,112}

\newcommand{\supptabind}{\hspace{0.65em}}
\newcommand{\suppgrp}[2]{\multicolumn{#1}{@{}l@{}}{\textit{#2}}}
\newcommand{\ci}[1]{\,{\footnotesize\textcolor{cicolor}{$#1$}}}
\newcommand{\gainonly}[1]{#1\phantom{\ci{[+0.000, +0.000]}}}
\newcommand{\hgain}[2]{\makebox[2.9em][r]{#1}\hspace{0.45em}\textcolor{slashcolor}{/}\,\makebox[2.85em][l]{#2}}
\newcommand{\hgaindag}[2]{\makebox[2.9em][r]{#1\rlap{$^\ddagger$}}\hspace{0.45em}\textcolor{slashcolor}{/}\,\makebox[2.85em][l]{#2}}
\newcommand{\hcost}[2]{\makebox[2.1em][r]{#1}\,\textcolor{slashcolor}{/}\,\makebox[2.45em][l]{#2}}
\newcommand{\hcorr}[2]{\makebox[2.55em][r]{$#1$}\,\textcolor{slashcolor}{/}\,\makebox[2.55em][l]{$#2$}}

\newtcolorbox{promptbox}[2][]{
  enhanced,
  colback=promptbg,
  colframe=promptborder,
  colbacktitle=prompthead,
  coltitle=prompttitle,
  fonttitle=\sffamily\bfseries\small,
  title={#2},
  boxrule=0.55pt,
  arc=2pt,
  left=11pt,
  right=11pt,
  top=7pt,
  bottom=7pt,
  toptitle=4pt,
  bottomtitle=4pt,
  #1
}

\makeatletter
\newcommand{\supplementlayout}{%
  \setlength{\intextsep}{9pt plus 3pt minus 2pt}%
  \captionsetup[table]{font=small,labelfont=bf,textfont=normalfont,skip=4.5pt}%
  \captionsetup[figure]{font=small,labelfont=bf,textfont=normalfont,skip=5.5pt}%
  \renewcommand\theadfont{\normalfont}%
  \renewcommand\cellset{\linespread{1.03}\selectfont}%
  \def\cvprsection{\@startsection {section}{1}{\z@}%
     {-10pt plus -2pt minus -2pt}{5pt} {\large\bf}}%
  \def\cvprsubsection{\@startsection {subsection}{2}{\z@}%
     {-8pt plus -2pt minus -1pt}{3.5pt} {\elvbf}}%
  \renewcommand{\paragraph}{\@startsection{paragraph}{4}{\z@}%
     {5.5pt plus 2pt minus 1pt}{-0.8em}{\normalsize\bfseries}}%
  \setlength{\@fptop}{0pt plus 0.38fil}%
  \setlength{\@fpsep}{14pt plus 0.14fil}%
  \setlength{\@fpbot}{0pt plus 0.62fil}%
  \setlength{\@dblfptop}{0pt plus 0.38fil}%
  \setlength{\@dblfpsep}{14pt plus 0.14fil}%
  \setlength{\@dblfpbot}{0pt plus 0.62fil}%
}
\makeatother

\definecolor{cvprblue}{rgb}{0.21,0.49,0.74}
\usepackage[pagebackref,breaklinks,colorlinks,allcolors=cvprblue]{hyperref}

\def\confName{CVPR}
\def\confYear{2027}

\title{\bench: Evaluating Multimodal Coding Agents on Interactive Fine-Grained 3D Head Reconstruction}

\author{%
\normalsize Haochen Zhao\textsuperscript{1}\thanks{Work done during an internship at Tencent.},
Yuanmeng Zhang\textsuperscript{2},
Dawei Zhu\textsuperscript{1},
Yifan Li\textsuperscript{3},
Zimo Meng\textsuperscript{1},
Zheng Li\textsuperscript{1}\\[3pt]
\normalsize Ruitao Liu\textsuperscript{4},
Bowen Qin\textsuperscript{2},
Yuhao Cheng\textsuperscript{2},
Haosheng Zou\textsuperscript{2},
Jeff Chen\textsuperscript{2}\thanks{Project lead.},
Sujian Li\textsuperscript{1}\thanks{Corresponding author.}\\[5pt]
\small \textsuperscript{1}Peking University
\quad \textsuperscript{2}Tencent
\quad \textsuperscript{3}Renmin University of China
\quad \textsuperscript{4}Tsinghua University
}
\hypersetup{
    pdftitle={BlenderFaceBench: Evaluating Multimodal Coding Agents on Interactive Fine-Grained 3D Head Reconstruction},
    pdfauthor={Haochen Zhao, Yuanmeng Zhang, Dawei Zhu, Yifan Li, Zimo Meng, Zheng Li, Ruitao Liu, Bowen Qin, Yuhao Cheng, Haosheng Zou, Jeff Chen, Sujian Li}
}

\makeatletter
\patchcmd{\@maketitle}{\begin{center}}{\vspace*{-40pt}\begin{center}}{}{}
\patchcmd{\@maketitle}{\vspace*{12pt}}{\vspace*{-19pt}}{}{}
\patchcmd{\@maketitle}{\vspace*{24pt}}{\vspace*{20pt}}{}{}
\makeatother

\begin{document}
\maketitle
\setlength{\stripsep}{7pt plus 2pt minus 2pt}
\begin{strip}
\centering
% Crop only the asset's outer white margins (left bottom right top); keep edge clearance.
\includegraphics[width=\textwidth,trim=0bp 10.25bp 0bp 22.75bp,clip]{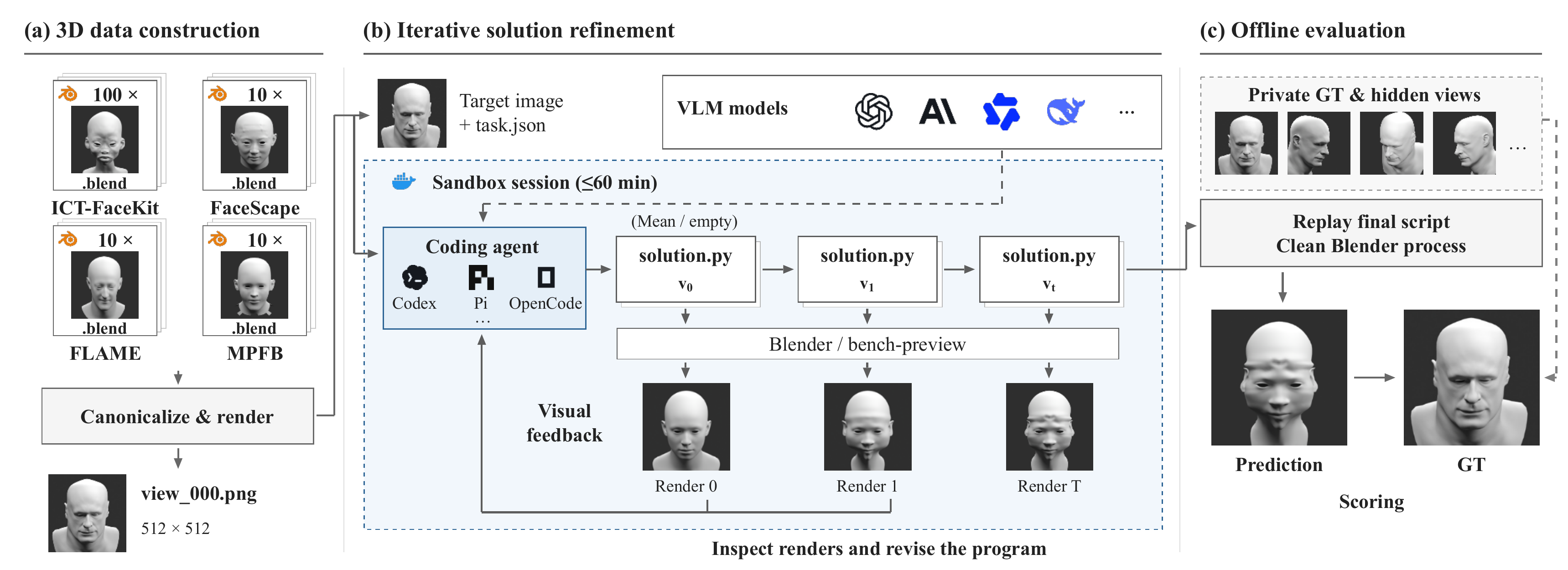}
\captionsetup{hypcap=false}
\captionof{figure}{\textbf{Overview of \bench.}
\textbf{(a)} Native \texttt{.blend} assets from four sources provide 130 canonicalized heads and single-view reference images.
\textbf{(b)} A VLM-powered coding agent iteratively refines a Blender script using self-rendered feedback in an isolated sandbox session.
\textbf{(c)} Clean replay enables canonical-space geometry evaluation against private GT, complemented by prior-relative scores and held-out-view silhouettes.}
\label{fig:teaser}
\captionsetup{hypcap=true}

\par\vspace{3pt}
\newsavebox{\projectglobe}
\sbox{\projectglobe}{%
  \tikz[x=1pt,y=1pt,baseline=-2.4pt,line width=0.65pt]{%
    \draw (0,0) circle[radius=4.3pt];
    \draw (0,0) ellipse[x radius=2pt,y radius=4.3pt];
    \draw (-4.3,0) -- (4.3,0);
  }%
}
\newcommand{\projectlink}[3]{%
  #2\hspace{4pt}\textbf{#3:}\enspace\href{#1}{\nolinkurl{#1}}%
}
\noindent\makebox[\linewidth][c]{%
  \normalfont\normalsize\urlstyle{same}%
  \renewcommand{\arraystretch}{0.94}%
  \begin{tabular}{@{}c@{}}
  \projectlink{https://blenderfacebench.github.io}{\usebox{\projectglobe}}{Homepage}\\
  % GitHub mark from https://github.githubassets.com/images/modules/logos_page/GitHub-Mark.png.
  \projectlink{https://github.com/BlenderFaceBench/BlenderFaceBench-Open}{%
    \raisebox{-0.22\height}{\includegraphics[height=8.5pt]{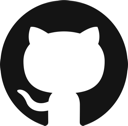}}%
  }{Code}\\
  % Hugging Face mark from https://huggingface.co/brand (hf-logo.png).
  \projectlink{https://huggingface.co/datasets/VectorZhao/BlenderFaceBench}{%
    \raisebox{-0.22\height}{\includegraphics[height=9.5pt]{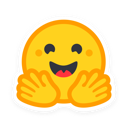}}%
  }{Data}
  \end{tabular}%
}\par
\end{strip}

\begin{abstract}
We introduce \bench, an interactive benchmark for evaluating the fine-grained 3D geometry reconstruction capabilities of multimodal coding agents. The benchmark contains 130 canonicalized heads across four data domains: ICT, FaceScape public, FLAME, and MPFB. Given a single frontal orthographic clay render, each task requires an agent to reconstruct a specific target head by writing Blender python code and inspecting multiview previews of its own output. This task requires agents to distinguish subtle craniofacial and facial-feature differences and translate visual judgments into continuous, executable geometric modifications. We provide mean-face and empty-scene initializations and evaluate improvements in target matching through offline replay of self-contained scripts. Across more than 2,100 containerized trials, we compare 10 configurations of 8 models and 4 harnesses. The strongest configuration, GPT-6 Astra (high), improves the strict F-score from the mean-face initialization of 0.229 to 0.321, leaving a substantial gap to a perfect score. On the same 20 ICT samples, only 3 of the 200 empty-scene runs across the 10 model configurations, each limited to 60 minutes, outperform the in-domain random-individual reference, revealing a capability gap in current models. Further analysis reveals distinct effects of initialization priors, reasoning effort, and local structural quality. Using face reconstruction as a testbed, \bench provides a reproducible platform for studying and improving target-driven, fine-grained 3D reconstruction by multimodal coding agents, laying a foundation for the next generation of multimodal coding agents with stronger geometric perception and coding capabilities.
\end{abstract}

\section{Introduction}
\label{sec:intro}
Multimodal coding agents are expanding from code completion and software repair to interaction with professional software~\cite{jimenez2024swebench,terminalbench2025}. In 3D modeling, Blender's Python interface enables agents to create geometry, inspect renders, and iteratively revise programs~\cite{du2024blenderllm,gu2024bladergym}. This capability opens a new path toward automated 3D reconstruction: agents could understand the target in an image, translate spatial judgments into executable geometric operations, and refine their predictions based on execution results. Evaluating this process helps establish the extent to which multimodal agents can perform target-driven 3D modeling.

The key question is therefore how to design a reconstruction task that reveals fine-grained geometric capabilities.
\textbf{Faces provide a discriminative testbed: they share the same category semantics, yet subtle geometric differences are highly salient to human observers.}
In general object or scene modeling (e.g., a chair or a room), discrete part presence and spatial arrangement---such as ``four legs with a backrest and armrests'' or ``a table in front of a chair''---largely determine the 3D structure.
In contrast, all faces share an identical part layout (two eyes above a central nose and a mouth below), where semantic decomposition loses discriminative power; individual appearance instead hinges on subtle, continuous variations in proportion and curvature (e.g., interocular distance, nasal bridge curvature, or jaw contour) that are visually salient yet difficult to specify concisely in language.
Face reconstruction thus requires capturing target-specific continuous geometry from an image and translating it into 3D modifications rather than generating a generic plausible face, while parametric head assets provide editable mean shapes and shared topology for controlled initializations and quantitative evaluation.
Although existing evaluations explore code-driven modeling (3D-GPT, SceneCraft, BlenderLLM) as well as procedural geometry, continuous deformation, and multiround verification (BlenderGym)~\cite{sun20233dgpt,hu2024scenecraft,du2024blenderllm,gu2024bladergym}, reconstructing a specific individual's head requires jointly examining whether predictions match subtle surface variations, exceed the mean-shape prior, and preserve valid geometric structure under local modifications.
Beyond post-registration shape error in conventional single-image face evaluation~\cite{sanyal2019learning,yang2020facescape}, matching quality for programmatic deliverables must also be assessed in the publicly specified canonical space.

\begin{figure*}[t]
% Crop only the asset's outer white margins; preserve the layout spacing.
\centering\includegraphics[width=\textwidth,trim=0bp 14.25bp 0bp 14.25bp,clip]{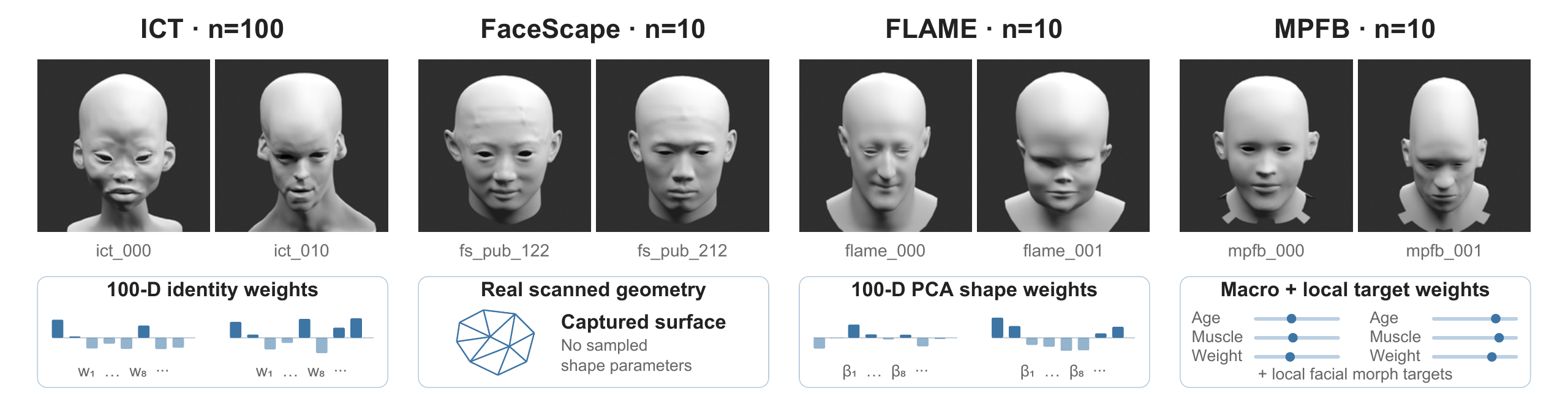}
\caption{Four data domains with example heads and selected construction parameters: identity weights (ICT), PCA shape weights (FLAME), and macro attributes with local morph targets (MPFB); FaceScape provides real scans. Construction parameters are not exposed to the agents.}
\label{fig:domains}
\end{figure*}

To this end, we introduce \textbf{\bench}, an interactive benchmark for fine-grained 3D head reconstruction by multimodal coding agents. The benchmark contains \textbf{100 ICT samples in the main set} and \textbf{10 cross-domain samples each} from FaceScape, FLAME, and MPFB, covering parametric shapes, real scans, statistical models, and character assets (Fig.~\ref{fig:domains}). All inputs are single frontal orthographic clay renders, controlling material, lighting, and framing. Within 60 minutes, agents write modeling programs and can freely inspect multiview previews of their own geometry before delivering a self-contained Blender script. The evaluator replays the script in a clean environment and scores the result (Fig.~\ref{fig:teaser}).

\bench focuses on two interconnected challenges. \textbf{First, visual diagnosis of subtle target differences.} A plausible generic head may resemble the reference in overall silhouette while missing local variations around the eyes and brows, nose, and mouth. We therefore report both whole-head and facial-region metrics and calibrate results against mean-face and in-domain random-individual scores. \textbf{Second, closed-loop execution from visual judgments to 3D modifications.} Models must select editing regions, control displacement and depth, and inspect the results from other viewpoints; even a correct diagnosis can be undermined by discontinuous mesh edits. Mean-face and empty-scene initializations further distinguish performance with and without editable geometry and canonical placement priors. Together, these challenges define the evaluation requirements from reference images to replayable geometric programs.

Across more than \textbf{2,100 containerized trials}, we evaluate 10 configurations of 8 models and 4 interaction harnesses, using specialized conventional computer vision methods---feed-forward face regression, general image-to-3D generation, and dedicated parametric fitting---as references. Our main findings are:
\begin{itemize}
    \item \textbf{Fine-grained refinement of a mean face is already feasible with state-of-the-art models.} The strongest configuration improves the strict F-score on the main set from the initialization of 0.229 to 0.321, with a facial Gain of $+0.200$; some models can nevertheless produce 3D faces with severe local holes or folds.
    \item \textbf{Initialization priors substantially affect results.} Only 3 of the 200 empty-scene runs on the same ICT20 outperform the in-domain random-individual reference. Cross-domain results further show that large improvements over the ICT initialization may reflect adjustments between head-shape domains rather than successful recovery of individual features. For a FaceScape target, for example, moving the initialization toward a typical FaceScape head shape can yield a large Gain while still missing the individual's distinctive facial features.
    \item \textbf{Reasoning effort and interaction configurations have different effects.} Increasing reasoning effort improves mean-face editing for both flagship models. Within each configuration, naturally occurring runtime and token consumption correlate weakly with Gain, and none of the four harnesses consistently dominates across models. These observations motivate further studies of fixed budgets and interaction mechanisms.
\end{itemize}

We hope \bench will support continued evaluation of fine-grained geometric capabilities in multimodal agents and provide a test environment for improving target-difference perception, geometric editing, and iterative verification. We will release the complete evaluation framework and data after the anonymous review period and support future expansion with additional data.

\section{Related Work}
\label{sec:related}
\paragraph{LLM agents for 3D modeling and reconstruction.}
Early efforts toward 3D modeling and reconstruction with LLM agents broadly fall into Blender-based and CAD-based approaches.
On the modeling side, Blender-based methods explore language-driven procedural generation and code self-improvement (3D-GPT, BlenderLLM)~\cite{sun20233dgpt,du2024blenderllm} or closed-loop planning, execution, and visual feedback (SceneCraft, VIGA, EZBlender)~\cite{hu2024scenecraft,yin2026viga,wang2026ezblender}; CAD-based methods learn executable parametric programs from text, point clouds, or preference feedback (Text2CAD, CAD-Recode, CADFusion)~\cite{khan2024text2cad,rukhovich2025cadrecode,wang2025cadfusion} or iteratively invoke CAD tools with execution and visual feedback (CAD-Assistant, IterCAD, ArtisanCAD)~\cite{mallis2025cadassistant,hu2026itercad,xu2026artisancad}.
As summarized in Table~\ref{tab:bench_comparison}, existing benchmarks cover only subsets of our evaluation dimensions: P3D-Bench~\cite{yang2026p3dbench} scores static parametric meshes without interactive visual loops; BlenderBench~\cite{yin2026viga} and BlenderWorld~\cite{fang2026blenderworld} support active self-viewing in interactive loops but omit 3D mesh evaluation; BlenderGym~\cite{gu2024bladergym} evaluates meshes in canonical space without active self-view or harness ablations; and 3DCodeBench~\cite{gao20263dcodebench} and 3DHarnessBench~\cite{liu20263dharnessbench} add 3D mesh scoring and harness comparisons without canonical-space placement checks.
Crucially, prior evaluations focus on discrete part structures or general scenes rather than continuous individual head-surface differences from a single image, and none controls initialization priors or calibrates improvements against in-domain individual references as \bench does.

\begin{table*}[t]
\centering
\caption{Comparison of \bench with existing programmatic 3D modeling and editing benchmarks.}
\label{tab:bench_comparison}
\small
\setlength{\tabcolsep}{5pt}
\begin{tabular}{lccccccc}
\toprule
Benchmark &
\shortstack{Interactive\\Visual Loop} &
\shortstack{Active\\Self-View} &
\shortstack{3D Mesh\\Evaluation} &
\shortstack{Canonical-Space\\Scoring} &
\shortstack{Harness\\Ablation} &
\shortstack{Controlled\\Init. Priors} &
\shortstack{Indiv. Surface \&\\Prior Calibration} \\
\midrule
P3D-Bench~\cite{yang2026p3dbench}          & \xmark & \xmark & \cmark & \xmark & \xmark & \xmark & \xmark \\
BlenderBench~\cite{yin2026viga}            & \cmark & \cmark & \xmark & \xmark & \xmark & \xmark & \xmark \\
BlenderWorld~\cite{fang2026blenderworld}   & \cmark & \cmark & \xmark & \xmark & \xmark & \xmark & \xmark \\
BlenderGym~\cite{gu2024bladergym}          & \cmark & \xmark & \cmark & \cmark & \xmark & \xmark & \xmark \\
3DCodeBench~\cite{gao20263dcodebench}      & \cmark & \cmark & \cmark & \xmark & \cmark & \xmark & \xmark \\
3DHarnessBench~\cite{liu20263dharnessbench}& \cmark & \cmark & \cmark & \xmark & \cmark & \xmark & \xmark \\
\midrule
\rowcolor{cgray}
\textbf{\bench (Ours)}                     & \cmark & \cmark & \cmark & \cmark & \cmark & \cmark & \cmark \\
\bottomrule
\end{tabular}
\end{table*}

\paragraph{Single-image face reconstruction and image-to-3D generation.}
3DMM, FLAME, ICT, and FaceScape provide face-shape priors or scan data~\cite{blanz1999morphable,li2017learning,kit2020ict,yang2020facescape}. DECA, EMOCA, MICA, and SMIRK estimate shape or expression from images~\cite{feng2021learning,danvevcek2022emoca,zielonka2022towards,retsinas2024smirk}, whereas LRM, TRELLIS, and Hunyuan3D-2 target general 3D generation~\cite{hong2023lrm,xiang2024structured,hunyuan3d22025}. We select DECA and Hunyuan3D-2 as references from different paradigms.

\paragraph{Geometric metrics and evaluation protocols.}
Category-mean shapes can obtain strong single-view reconstruction scores~\cite{tatarchenko2019single}. Strict F-score~\cite{knapitsch2017tanks} and facial regions help reveal local differences but can miss small-area defects. Compared with post-registration evaluation~\cite{sanyal2019learning}, our canonical-space scoring retains placement error and calibrates results using in-domain references.

\section{Benchmark and Evaluation Protocol}
\label{sec:method}
\subsection{Task and Replayable Deliverables}
\label{sec:task_def}

We build our evaluation on Harbor framework~\cite{shi2026harboradapters}. 

Each run takes place in an isolated container with 8 CPUs, 16 GB of memory, and no GPU, with a 60-minute wall-clock limit, a 128k context window, and at most 32,768 output tokens per response. The main harness, \texttt{terminus-2-vision}, adds image inspection to Harbor's minimal \texttt{terminus-2} harness. We also compare Codex CLI~\cite{openai2026codexcli}, OpenCode~\cite{anomaly2026opencode}, and Hermes Agent~\cite{nous2026hermesagent} on a fixed ICT20 subset. All four support file editing, command execution, and preview inspection. The preview tool accepts specified yaw and pitch angles; it returns only renders of the current prediction, without ground-truth feedback.

For each test case, the input is a single $512\times512$ frontal orthographic clay render together with camera parameters and task limits. Heads exclude hair, eyeballs, teeth, and accessories, with standardized material and lighting. Agents cannot access the ground-truth mesh or target reference images from other viewpoints, but they can render their own geometry. The public coordinate convention is rightward $+X$, upward $+Z$, and forward-facing $-Y$. Crown-to-chin height is 2, the origin is their midpoint, and the scoring region is $z\ge-1.05$. The camera's orthographic scale is 2.6.

The \texttt{solution.py} file on disk when the session ends is the final answer. The evaluator replays it in a fresh, network-isolated Blender 5.2.1~\cite{blender2026pythonapi} process, without external mesh files, within 120 seconds and with at most 500,000 triangles. Missing, failed, or over-limit submissions receive zero scores; infrastructure exceptions are recorded separately (supplementary material). The oracle self-check using the ground-truth construction script returns 1.000 and is used to check the delivery and scoring pipeline.

\subsection{Data and Initializations}
\label{sec:dataset}
The main set contains 100 parametric heads from ICT-FaceKit~\cite{kit2020ict}, selected from a candidate pool by maximin sampling to increase shape diversity. Each additional domain contains 10 samples: real FaceScape scans~\cite{yang2020facescape}, statistical FLAME heads~\cite{li2017learning}, and MPFB 2.0.17 character assets~\cite{makehuman2026mpfb2017}. All ground-truth heads are canonicalized in advance using five landmarks: the two outer eye corners, nose tip, chin, and crown. These sources combine the realism of scanned faces with the precision and diversity of geometrically constructed assets, making them suitable for benchmark construction.

\textbf{Easy} provides an editable ICT mean-face script. For ICT targets, the initialization is generated from the mean shape parameters of the other 99 samples, excluding the target from its own initialization. The three additional domains use the mean face of all 100 ICT samples. \textbf{Default} starts from an empty scene. These settings progress from easier to harder tasks, enabling a more fine-grained assessment of geometric editing and construction capabilities.

\begin{table}[t]
\centering
\caption{Data and reference scores (F@0.5\%). The in-domain mean is the leave-one-out mean face; the random-individual reference averages matching scores against the other individuals in the same domain.}
\label{tab:ref_lines}
\small\setlength{\tabcolsep}{3.5pt}
\begin{tabular}{lrrrr}
\toprule
Source & $n$ & ICT init. & \shortstack{In-domain\\mean} & \shortstack{Random\\individual}\\
\midrule
ICT &100&0.229&0.229&0.179\\
FaceScape &10&0.191&0.518&0.424\\
FLAME &10&0.217&0.218&0.184\\
MPFB &10&0.169&0.209&0.164\\
\bottomrule
\end{tabular}
\end{table}

\subsection{Metrics and Limits of Interpretation}
\label{sec:metrics}
The evaluator samples 100,000 points from each clipped surface in proportion to area, approximates surface distances using bidirectional nearest sampled points, and computes the harmonic mean of precision and recall. Thresholds are 0.5\% and 2\% of the ground-truth bounding-box diagonal, denoted by \fstrict and \floose.

For each sample, we define the following using its initialization score $\initf$ and in-domain random-individual reference score $\otherf$:
\begin{equation}
\gain=\frac{F-\initf}{1-\initf},\qquad
\ident=\frac{F-\otherf}{1-\otherf}.
\label{eq:gain_identity}
\end{equation}
Reported values are computed per sample and then averaged. Gain measures the fraction of remaining F-score improvement achieved; Identity indicates only a geometric advantage over an in-domain random individual. In particular, the ICT mean face itself has an Identity of $+0.062$, so a positive score alone does not demonstrate recovery of target-specific features. The target-domain mean face is also a necessary reference (Table~\ref{tab:ref_lines}).

Auxiliary metrics include silhouette IoU over 10 held-out viewpoints, F-scores for five regions---brows and eyes, nose, mouth, jaw, and ears---as well as combined facial Gain and angular normal error. Regions are defined using ground-truth landmarks; predicted samples are assigned to the region of their nearest ground-truth point. These metrics reduce dilution by the large cranial surface and provide a more comprehensive analysis.

We report $95\%$ t confidence intervals for means over samples. Paired comparisons use the Wilcoxon signed-rank test with Holm correction within each comparison family, and overall model comparisons use the Friedman test.

\section{Experiments and Analysis}
\label{sec:experiments}
\subsection{Models and Reference Methods}
\label{sec:exp_setup}
Evaluation includes three self-hosted models, Qwen3.8-27B~\cite{qwen2026qwen38card}, GLM-5.3-Flash~\cite{zai2026glm53flashcard}, and DeepSeek-V4.1-Flash-1017~\cite{deepseek2026v41flash1017card}, and five API models, GPT-6 Luna~\cite{openai2026gpt6luna}, Claude Sonnet 5.5~\cite{anthropic2026sonnet55}, Gemini 3.5 Flash-Lite~\cite{google2026gemini35flashlite}, Claude Opus 5.5~\cite{anthropic2026opus55}, and GPT-6 Astra~\cite{openai2026gpt6astra}. For Claude Opus 5.5 and GPT-6 Astra, we test both the lowest accepted effort (none/low, $^\dagger$) and high, yielding 10 valid configurations. Other configurations use high reasoning settings accepted by their interfaces. Self-hosted models run on an H20 cluster.

Each configuration contains 100 ICT main-set runs, 30 cross-domain runs, 20 empty-scene runs, and three groups of 20 runs with additional harnesses, totaling 210 runs per configuration and 2,100 across the 10 configurations, plus diagnostics and reruns. The main experiment uses Easy and \texttt{terminus-2-vision}; initialization and harness comparisons use the fixed ICT20. All geometric quality comparisons use the same evaluator.

Non-agent references include DECA~\cite{feng2021learning} and Hunyuan3D-2~\cite{hunyuan3d22025}. DECA regresses FLAME face parameters from a single image and represents expression-dependent facial detail with UV displacement maps. We use only the neutral coarse mesh generated from its identity shape parameters and canonicalize it using landmarks on the predicted mesh itself. Hunyuan3D-2 conditions on an image and generates 3D shapes in the ShapeVAE latent space using a diffusion Transformer trained with flow matching. We use only its shape-generation stage, determine scale and horizontal and vertical position from the input silhouette, and use the mean head to determine front-to-back translation. Easy agents receive an editable mean face and an interaction budget, so, strictly speaking, these two methods can also be compared with Default runs.

We also include dedicated frontal-image fitting with a known ICT shape basis (F of 0.297) as a reference. The per-sample best envelope over 10 configurations takes the highest score for each target and reflects only complementarity among tested configurations.
\begin{table*}[t]
\centering
\caption{Core metrics for Easy (ICT100) and Default (ICT20, empty scene). Gain is reported as mean and 95\% CI; Identity uses the in-domain random-individual reference. The two sets differ in sample scope; paired initialization comparisons appear in supplementary Table~\ref{tab:s_start}. Reference rows are aggregated over the cases in each column and do not denote rerunning from an empty scene.}
\label{tab:main_results}
{\small
\setlength{\tabcolsep}{2pt}
\begin{tabular}{lcccccccc}
\toprule
& \multicolumn{5}{c}{Easy (ICT100)} & \multicolumn{3}{c}{Default (ICT20)} \\
\cmidrule(lr){2-6}\cmidrule(lr){7-9}
Method & \fstrict & \gain [95\% CI] & \ident & Facial F & Facial Gain & \fstrict & \ident & Facial F \\
\midrule
\textit{Mean-face initialization} & \textit{0.229} & \textit{0.000} & \textit{+0.062} & \textit{0.262} & \textit{0.000} & 0.226 & +0.061 & 0.279 \\
\midrule
\textbf{DECA} & 0.219 & $-$0.016 $[-0.028, -0.003]$ & +0.050 & 0.256 & $-$0.012 & 0.206 & +0.036 & 0.274 \\
\textbf{Hunyuan3D-2} & 0.190 & $-$0.057 $[-0.076, -0.038]$ & +0.014 & 0.177 & $-$0.129 & 0.174 & $-$0.004 & 0.163 \\
\textit{ICT 3DMM frontal fitting} & \textit{0.297} & +0.084 $[+0.063, +0.106]$ & \textit{+0.144} & 0.303 & +0.045 & 0.292 & +0.140 & 0.294 \\
\midrule
\textbf{GPT-6 Astra (high)}$^*$ & \textbf{0.321} & \textbf{+0.118} $[+0.099, +0.137]$ & \textbf{+0.174} & \textbf{0.411} & \textbf{+0.200} & 0.084 & $-$0.113 & 0.132 \\
\textbf{Claude Opus 5.5 (high)} & 0.297 & +0.087 $[+0.073, +0.102]$ & +0.145 & 0.351 & +0.115 & 0.078 & $-$0.120 & 0.089 \\
\textbf{GPT-6 Astra (low)}$^\dagger$ & 0.281 & +0.065 $[+0.050, +0.081]$ & +0.126 & 0.358 & +0.126 & \textbf{0.103} & \textbf{$-$0.090} & \textbf{0.139} \\
\textbf{Claude Opus 5.5 (none)}$^\dagger$ & 0.277 & +0.061 $[+0.048, +0.074]$ & +0.120 & 0.316 & +0.070 & 0.094 & $-$0.101 & 0.116 \\
\textbf{Claude Sonnet 5.5} & 0.277 & +0.061 $[+0.048, +0.074]$ & +0.121 & 0.317 & +0.070 & 0.097 & $-$0.097 & 0.103 \\
\textbf{GPT-6 Luna} & 0.268 & +0.049 $[+0.037, +0.060]$ & +0.109 & 0.309 & +0.060 & 0.082 & $-$0.115 & 0.113 \\
\textbf{GLM-5.3-Flash} & 0.251 & +0.026 $[+0.009, +0.043]$ & +0.088 & 0.276 & +0.013 & 0.067 & $-$0.134 & 0.086 \\
\textbf{Qwen3.8-27B} & 0.245 & +0.019 $[+0.004, +0.035]$ & +0.082 & 0.259 & $-$0.011 & 0.065 & $-$0.136 & 0.095 \\
\textbf{Gemini 3.5 Flash-Lite}$^*$ & 0.182 & $-$0.064 $[-0.085, -0.043]$ & +0.005 & 0.208 & $-$0.080 & 0.054 & $-$0.149 & 0.027 \\
\textbf{\begin{tabular}[t]{@{}l@{}}DeepSeek-V4.1-\\Flash-1017\end{tabular}} & 0.258 & +0.034 $[+0.017, +0.051]$ & +0.097 & 0.277 & +0.011 & 0.055 & $-$0.147 & 0.073 \\
\bottomrule
\end{tabular}}
\end{table*}

\subsection{Improving Target Matching from a Mean Face}
\label{sec:main_exp}
Table~\ref{tab:main_results} shows positive mean Gain for most configurations (see also supplementary Fig.~\ref{fig:evidence}a). GPT-6 Astra (high) improves \fstrict from 0.229 at initialization to 0.321, with Gain $+0.118$ and facial Gain $+0.200$; Claude Opus 5.5 (high) achieves 0.297 and $+0.087$. The mean score of GPT-6 Astra (high) exceeds the ICT frontal-fitting reference in this experiment; this comparison uses a mean-face geometric prior.

Within-model effort comparisons show a mean Gain increase of $0.053$ from low to high for GPT-6 Astra (better on 78/100 cases) and $0.026$ from none to high for Claude Opus 5.5 (better on 73/100 cases), both significant after Holm correction. Complete regional results show that whole-head and facial rankings do not fully agree: GPT-6 Astra (low) has a facial F of 0.358, above Claude Opus 5.5 (high) at 0.351, while Qwen3.8-27B has positive whole-head Gain but negative facial Gain. Whole-head means therefore cannot replace local geometric evaluation.

Under the same canonical-space scoring, DECA and Hunyuan3D-2 achieve F-scores of 0.219 and 0.190, respectively.

\subsection{Initialization, Harnesses, and Cross-Domain Evaluation}
\label{sec:exp_ablations}
\paragraph{Initialization has a substantial effect.}
In the paired ICT20 comparison (supplementary Table~\ref{tab:s_start} and Fig.~\ref{fig:evidence}b), only three of the 200 empty-scene runs have positive Identity; mean \fstrict ranges from 0.054--0.103 across configurations. Mean Identity is $0.153$--$0.288$ higher under Easy than Default. This shows strong dependence on an editable canonical geometric initialization, although low zero-registration scores alone do not imply that no shape was recovered. In the empty-scene setting, high reasoning effort has lower mean Identity than the corresponding minimum effort; the gains observed for mean-face editing do not transfer (complete tables in the supplementary material).

\begin{figure*}[t]
\centering
% The generated asset already has tight outer bounds.
\includegraphics[width=\textwidth]{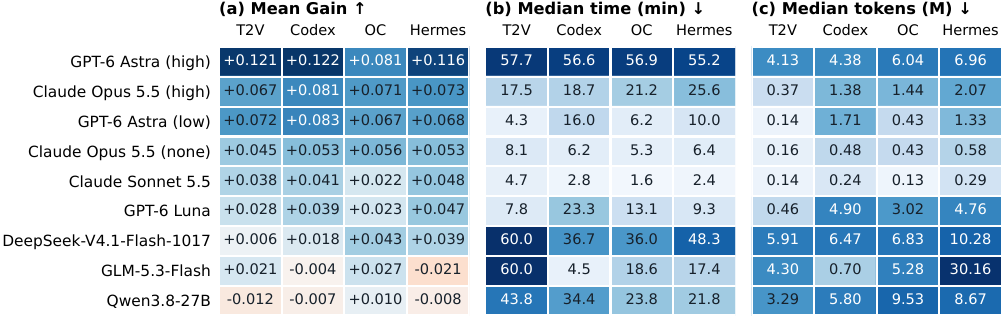}
\caption{\textbf{Harness quality and interaction cost on the same ICT20 targets.}
T2V denotes \texttt{terminus-2-vision}; OC denotes OpenCode. Cells show mean whole-head Gain, median wall-clock minutes, and median cumulative input/output tokens (millions, including repeated context). Colors are scaled separately by panel; token colors use a logarithmic scale. Gemini is omitted because its Codex/Hermes integrations were invalid. Quality rankings vary by model, while time and token costs can differ substantially. Complete quality and cost results appear in supplementary Tables~\ref{tab:s_harness}--\ref{tab:s_effort}.}
\label{fig:harness_quality_cost}
\end{figure*}

\paragraph{Harnesses affect cost; performance effects vary by model.}
Figure~\ref{fig:harness_quality_cost} compares quality and interaction cost on the same ICT20 targets. None of the four harnesses consistently dominates across models. The pooled descriptive test for the nine configurations valid under all four harnesses gives $p=0.38$, while configuration-specific tests for GPT-6 Astra (high) and GLM-5.3-Flash give $p=0.017$ and $0.007$. Switching Qwen3.8-27B from \texttt{terminus-2-vision} to OpenCode increases median cumulative tokens from 3.29M to 9.53M while reducing median runtime from 43.8 to 23.8 minutes (supplementary Tables~\ref{tab:s_harness}--\ref{tab:s_harness_cost}).

\paragraph{Cross-domain improvement does not imply individual recovery.}
On FaceScape, Claude Opus 5.5 (high) reaches Gain $+0.302$, but Identity is only $+0.016$, with F of 0.438, still below the in-domain leave-one-out mean face at 0.518. Large gains may primarily reflect adjustments between head-shape domains and do not establish recovery of individual features. On FLAME, GPT-6 Astra (high) has Gain $+0.089$, above DECA at $+0.062$; on MPFB, Hunyuan3D-2 has whole-head Gain $+0.134$, above all agents, but facial Gain is $-0.004$. Each domain contains only 10 samples, and the initialization is always the ICT mean face, so these rankings are exploratory.

% Declare the full-width trajectory figure early so it stays near the failure analysis.
\begin{figure*}[t]
\centering
% Crop only the asset's outer white margins; preserve the layout spacing.
\includegraphics[width=0.94\textwidth,trim=0bp 8bp 0bp 13bp,clip]{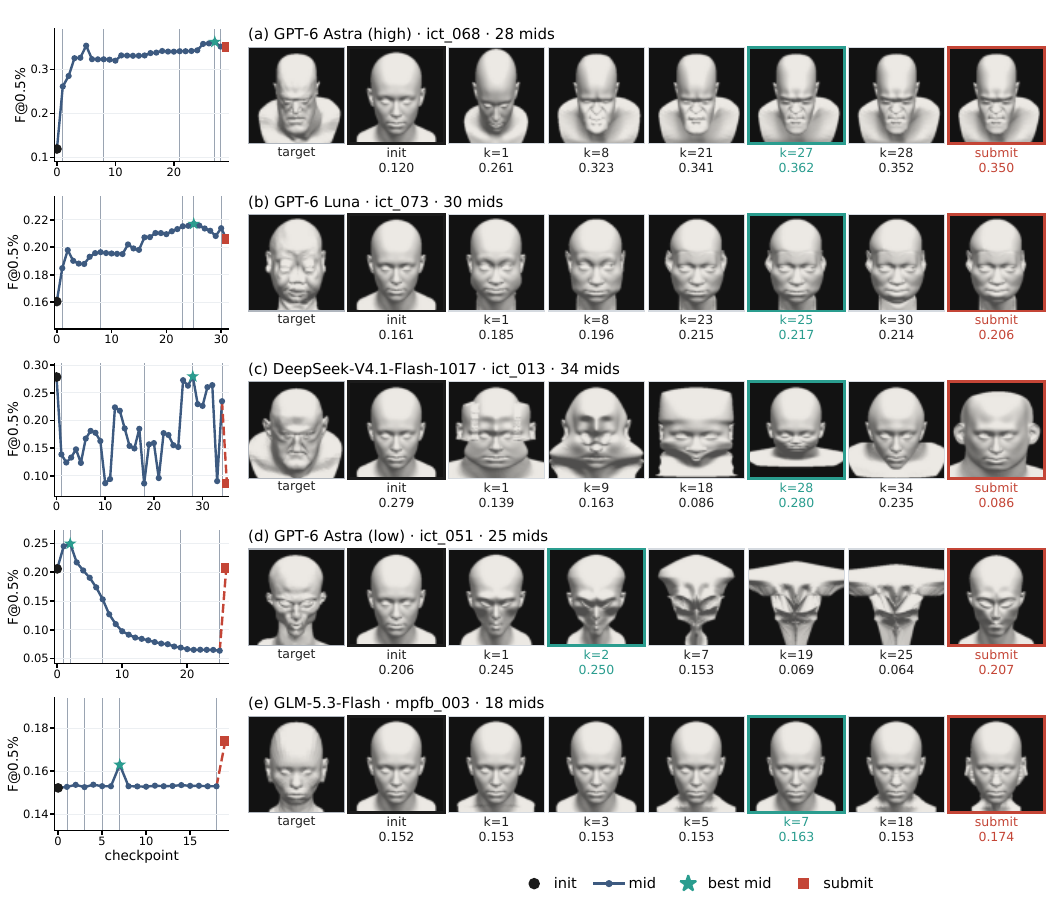}
\caption{Selected reconstruction trajectories with post-hoc F@0.5\% scores and replayed renders. Checkpoints index saved intermediate scripts; stars mark the best intermediate score and squares the final submission. Axes use different ranges, and GT scores are never provided to the agents.}
\label{fig:long_traj}
\end{figure*}

\subsection{Interaction Cost and Reasoning Effort}
\label{sec:analysis}
\label{sec:scaling_analysis}
Supplementary Table~\ref{tab:s_effort} reports the complete main-set interaction costs, termination states, and per-sample correlations. Within each of the 10 configurations, the Spearman correlations of cumulative tokens and wall-clock time with Gain satisfy $|\rho|\le0.21$. The three self-hosted models have median runtimes of 44--60 minutes and cumulative token counts of 3.53--6.10M, without achieving higher mean scores as a result.

Increasing GPT-6 Astra effort from low to high raises median runtime from 4.1 to 57.5 minutes and tokens from 0.13M to 4.15M; increasing Claude Opus 5.5 effort from none to high raises runtime from 8.4 to 16.7 minutes and tokens from 0.18M to 0.39M. Both improve mean-face editing.

\paragraph{Termination states.}
GLM-5.3-Flash reaches the time limit on 73/100 main-set runs, and 27 final surface scores equal the initialization score. Equal scores can correspond to unchanged or different geometry; final scores alone cannot establish the editing process. The supplementary material reports complete termination states; the next section analyzes geometric editing and final delivery through replays of intermediate states from selected trajectories.

% Place the pairwise table near the visual matching analysis without stacking both large floats.
\begin{table*}[t]
\centering
\caption{Offline pairwise visual matching on 45 candidate pairs, each repeated three times with both presentation orders. F/H denotes F-score/human reference; strict accuracy requires both orders to select the correct candidate. Rates are percentages; $\tau_a$ uses aggregate win counts. Correlated repetitions preclude treating all 135 groups as independent for confidence intervals.}
\label{tab:visual_pairwise}
{\small
\setlength{\tabcolsep}{2.5pt}
\begin{tabular}{lccccccc}
\toprule
Model & \shortstack{Strict acc.\\F-score} & \shortstack{Strict acc.\\Human} & \shortstack{AB/BA\\consistency} & \shortstack{Consistent wrong\\F/H} & \shortstack{Per-query acc.\\F/H} & Choose A & \shortstack{Kendall $\tau_a$\\F/H} \\
\midrule
GLM-5.3-Flash & 88.9 (120/135) & 86.7 (117/135) & 91.9 & 3.0 / 5.2 & 93.0 / 90.7 & 51.1 & 0.82 / 0.82 \\
Qwen3.8-27B & 84.4 (114/135) & 83.0 (112/135) & 85.2 & 0.7 / 2.2 & 91.9 / 90.4 & 46.3 & 0.91 / 0.82 \\
\shortstack[l]{DeepSeek-V4.1-\\Flash-1017} & 78.5 (106/135) & 76.3 (103/135) & 83.7 & 5.2 / 7.4 & 86.7 / 84.4 & 42.8 & 0.82 / 0.73 \\
\bottomrule
\end{tabular}}
\end{table*}

\section{Failure Modes and Iterative Analysis}
\label{sec:failure_modes}

We decompose the capabilities required for multimodal coding agents to reconstruct 3D faces into three components: visual difference perception, geometric editing execution, and result evaluation and selection. Figure~\ref{fig:long_traj} juxtaposes replay scores and renders from several representative long-horizon trajectories, providing a basis for analysis.

\subsection{Visual Difference Perception}
\label{sec:failure_perception}
\paragraph{Offline pairwise matching.}
We design a simple experiment to test models' perception of facial similarity independently. Given a target reference and two anonymous candidate reconstructions, the model selects the candidate closer to the target. This isolates code generation and historical version management to test visual matching judgments over given candidates. The 10 selected candidates form 45 pairs; each pair is queried in both AB and BA orders, repeated three times, yielding 135 paired tests and 270 queries. After reversing presentation order, responses are mapped back to candidate identities: a pair is strictly correct only when both responses select the same candidate that is better under the reference ranking. Order-inconsistent pairs remain in the denominator and count as errors. We additionally report consistent-but-wrong responses, per-query accuracy, and the proportion selecting label A. We aggregate each candidate's selections across all valid responses as its win count (at most 54 wins per candidate), then compute Kendall $\tau_a$ against the F-score or human ranking. Candidate pairs tied in win count contribute 0, with a fixed denominator of 45.

The human and \fstrict rankings agree on 43 of the 45 pairs (95.6\%), supporting correspondence between visible matching quality and geometric metrics within this candidate set. Table~\ref{tab:visual_pairwise} reports results against both references. Strict F-score accuracy is 88.9\%, 84.4\%, and 78.5\% for GLM-5.3-Flash, Qwen3.8-27B, and DeepSeek-V4.1-Flash-1017, respectively; the corresponding human-reference accuracies are 86.7\%, 83.0\%, and 76.3\%.

Qwen3.8-27B achieves a $\tau_a$ of 0.91 against F-score for its aggregate ranking, but its order consistency is lower than that of GLM-5.3-Flash. Models already exhibit some visual matching ability when given candidates, but this does not guarantee localization of all differences, effective modifications, or retention of results over long trajectories.

\subsection{Geometric Editing Execution}
\label{sec:failure_editing}
Geometric editing requires translating visual diagnosis into effective region selection, deformation, and edit magnitude. Correctly identifying a difference does not guarantee sufficient modification of the target geometry.

\paragraph{Local diagnosis does not translate into effective deformation.}
On \texttt{mpfb\_003}, GLM-5.3-Flash identifies differences in ear and neck width but mainly adjusts the silhouette half-width of the ICT mean face (Fig.~\ref{fig:long_traj}e). Across 18 intermediate states, mean vertex displacement from the initialization is only about 0.001--0.003, with almost no facial movement, and \fstrict mostly remains near 0.153. A variant that modifies the ears reaches 0.163 before returning to about 0.153. Retaining the ICT topology is consistent with Easy; the issue is insufficient modification of target-relevant regions.

The final submission reduces the half-width control from $\pm1.00$ to $\pm0.80$. The ears and neck move, but mean facial displacement remains only about 0.003. Mean shoulder and neck displacement outside the ROI is 0.135, compared with 0.016 inside, and the final \fstrict is 0.174. This supports a diagnosis of insufficient editing extent and magnitude: multiple operations mainly change the silhouette, shoulders, and neck without sufficiently modifying the facial surface toward the target.

GPT-6 Luna on \texttt{ict\_073} instead shows overall improvement (Fig.~\ref{fig:long_traj}b): \fstrict rises from 0.161 at initialization to 0.196 at the 8th state and 0.217 at the 25th, ending at 0.206. Although not strictly monotonic, this case shows that successive edits can accumulate improvements in target matching.

\paragraph{Executable modifications can still damage local structure.}
The mouth hole and transverse folds in supplementary Fig.~\ref{fig:qualitative_traj} show that replayable scripts can still introduce local damage, and positive Gain does not guarantee structural validity. In Fig.~\ref{fig:long_traj}d, GPT-6 Astra (low) on \texttt{ict\_051} reaches a \fstrict of 0.250 at the 2nd state, then drops to 0.153, 0.069, and 0.064 at the 7th, 19th, and 25th states, accompanied by facial stretching and collapse. Iterative editing can thus exacerbate local damage.

\subsection{Result Evaluation and Selection}
\label{sec:failure_selection}
This component requires agents to evaluate candidate states and preserve existing progress while continuing to edit, save, restore, and submit. For the set $\mathcal C$ consisting of saved initial, intermediate, and final states, we use
\begin{equation}
\Delta_{\mathrm{sel}}=\max_{c\in\mathcal C}F(c)-F(c_{\mathrm{final}})
\label{eq:selection_gap}
\end{equation}
to describe the gap between the final score and the highest score among saved candidates, where $F$ is the strict F-score and the quantity is defined by post-hoc ground-truth evaluation.

\paragraph{A better intermediate state is not retained in the final submission.}
DeepSeek-V4.1-Flash-1017 exhibits strongly non-monotonic changes on \texttt{ict\_013} (Fig.~\ref{fig:long_traj}c): the initialization scores 0.279, the 28th intermediate state reaches 0.280, the 34th scores 0.235, and the final submission drops to 0.086, giving $\Delta_{\mathrm{sel}}=0.194$. This run fails to preserve a higher-scoring intermediate state in its final deliverable, potentially reflecting issues in candidate evaluation, version retention or restoration, or pre-submission verification.

For the same GPT-6 Astra (low) case, the final \fstrict recovers to 0.207, above the late-stage 0.064 and close to the initialization's 0.206 (Fig.~\ref{fig:long_traj}d). Although still below the early best of 0.250 ($\Delta_{\mathrm{sel}}=0.043$), it nevertheless shows the effect of retaining or restoring a more reliable version, indicating some ability of GPT-6 Astra (low) to select intermediate states and roll back.

\section{Conclusion}
\label{sec:conclusion}
We introduce \bench, which uses head reconstruction with shared category semantics and subtle individual differences to evaluate fine-grained geometric capabilities in multimodal coding agents. Standardized imaging, controlled initializations, replayable programs, and multilevel evaluation connect the perception of target differences to 3D operations, complementing general 3D generation and editing benchmarks.

\clearpage
{
    \small
    \bibliographystyle{ieeenat_fullname}
    \bibliography{main}

@String(CVPR  = {CVPR})

@String(ICCV  = {ICCV})

@String(ECCV  = {ECCV})

@String(NIPS  = {NeurIPS})

@String(ICLR  = {ICLR})

@String(TOG   = {ACM TOG})

@article{shi2026harboradapters,
  author = {Shi, Lin and Lin, Haowei and Zhu, Zixuan and others},
  title = {{Harbor Adapters and Harbor-Index}: Infrastructure and a Curated Meta-Dataset for Large-Scale Agentic Evaluation},
  journal = {arXiv preprint arXiv:2609.04298},
  year = {2026},
  url = {https://arxiv.org/abs/2609.04298}
}

@misc{qwen2026qwen38card,
  author = {{Qwen Team}},
  title = {{Qwen3.8-27B}},
  howpublished = {Official model card. \url{https://huggingface.co/Qwen/Qwen3.8-27B}},
  url = {https://huggingface.co/Qwen/Qwen3.8-27B},
  year = {2026},
  note = {Accessed 2026-11-05}
}

@misc{zai2026glm53flashcard,
  author = {{Z.ai}},
  title = {{GLM-5.3-Flash}},
  howpublished = {Official model card. \url{https://huggingface.co/zai-org/GLM-5.3-Flash}},
  url = {https://huggingface.co/zai-org/GLM-5.3-Flash},
  year = {2026},
  note = {Accessed 2026-11-05}
}

@misc{deepseek2026v41flash1017card,
  author = {{DeepSeek-AI}},
  title = {{DeepSeek-V4.1-Flash-1017}},
  howpublished = {Official model card. \url{https://huggingface.co/deepseek-ai/DeepSeek-V4.1-Flash-1017}},
  url = {https://huggingface.co/deepseek-ai/DeepSeek-V4.1-Flash-1017},
  year = {2026},
  note = {Accessed 2026-11-05}
}

@misc{openai2026gpt6luna,
  author = {{OpenAI}},
  title = {{GPT-6 Luna}},
  howpublished = {Official model documentation. \url{https://developers.openai.com/api/docs/models/gpt-6-luna}},
  url = {https://developers.openai.com/api/docs/models/gpt-6-luna},
  year = {2026},
  note = {Accessed 2026-11-05}
}

@misc{openai2026gpt6astra,
  author = {{OpenAI}},
  title = {{GPT-6 Astra}},
  howpublished = {Official model documentation. \url{https://developers.openai.com/api/docs/models/gpt-6-astra}},
  url = {https://developers.openai.com/api/docs/models/gpt-6-astra},
  year = {2026},
  note = {Accessed 2026-11-05}
}

@misc{anthropic2026sonnet55,
  author = {{Anthropic}},
  title = {Introducing {Claude Sonnet 5.5}},
  howpublished = {Official release announcement. \url{https://www.anthropic.com/claude-sonnet-5-5}},
  url = {https://www.anthropic.com/claude-sonnet-5-5},
  month = sep,
  year = {2026},
  note = {Published September 28, 2026; accessed 2026-11-05}
}

@misc{anthropic2026opus55,
  author = {{Anthropic}},
  title = {Introducing {Claude Opus 5.5}},
  howpublished = {Official release announcement. \url{https://www.anthropic.com/claude-opus-5-5}},
  url = {https://www.anthropic.com/claude-opus-5-5},
  month = sep,
  year = {2026},
  note = {Published September 22, 2026; accessed 2026-11-05}
}

@misc{google2026gemini35flashlite,
  author = {{Google DeepMind}},
  title = {{Gemini 3.5 Flash-Lite} Model Card},
  howpublished = {\url{https://deepmind.google/models/model-cards/gemini-3-5-flash-lite/}},
  url = {https://deepmind.google/models/model-cards/gemini-3-5-flash-lite/},
  month = jul,
  year = {2026},
  note = {Published July 21, 2026; accessed 2026-11-05}
}

@misc{makehuman2026mpfb2017,
  author = {{MakeHuman Community}},
  title = {{MPFB 2.0.17}},
  howpublished = {Official release notes. \url{https://static.makehumancommunity.org/mpfb/releases/release_2017.html}},
  url = {https://static.makehumancommunity.org/mpfb/releases/release_2017.html},
  month = jul,
  year = {2026},
  note = {Released July 22, 2026; accessed 2026-11-05}
}

@misc{blender2026pythonapi,
  author = {{Blender Foundation}},
  title = {{Blender 5.2 Python API}: {API} Overview},
  howpublished = {Official documentation. \url{https://docs.blender.org/api/5.2/info_overview.html}},
  url = {https://docs.blender.org/api/5.2/info_overview.html},
  year = {2026},
  note = {Accessed 2026-11-05}
}

@article{yin2026viga,
  author = {Yin, Shaofeng and Ge, Jiaxin and Wang, Zora Zhiruo and Wang, Chenyang and Li, Xiuyu and Black, Michael J. and Darrell, Trevor and Kanazawa, Angjoo and Feng, Haiwen},
  title = {Vision-as-Inverse-Graphics Agent via Interleaved Multimodal Reasoning},
  journal = {arXiv preprint arXiv:2601.11109},
  year = {2026},
  url = {https://arxiv.org/abs/2601.11109}
}

@article{wang2026ezblender,
  author = {Wang, Hao and Zhu, Wenhui and Tang, Shao and Wang, Zhipeng and Dong, Xuanzhao and Li, Xin and Chen, Xiwen and Bastola, Ashish and Huang, Xinhao and Wang, Yalin and Razi, Abolfazl},
  title = {{EZBlender}: Efficient {3D} Editing with Plan-and-{ReAct} Agent},
  journal = {arXiv preprint arXiv:2601.07143},
  year = {2026},
  url = {https://arxiv.org/abs/2601.07143}
}

@article{gao20263dcodebench,
  author = {Gao, Yipeng and Shu, Lei and Ye, Genzhi and Xiong, Xi and Makadia, Ameesh and Guo, Meiqi and Itti, Laurent and Chen, Jindong},
  title = {{3DCodeBench}: Benchmarking Agentic Procedural {3D} Modeling Via Code},
  journal = {arXiv preprint arXiv:2606.01057},
  year = {2026},
  url = {https://arxiv.org/abs/2606.01057}
}

@article{liu20263dharnessbench,
  author = {Liu, Ling and Gong, Bingchen and Parakkat, Amal Dev and Ovsjanikov, Maks},
  title = {{3DHarnessBench}: Probing Agentic {3D}-to-Code Capabilities of Frontier Vision-Language Models},
  journal = {arXiv preprint arXiv:2609.06535},
  year = {2026},
  url = {https://arxiv.org/abs/2609.06535}
}

@article{fang2026blenderworld,
  author = {Fang, Xinyu and Wang, Yuchi and Yu, Weihao and Fang, Rongyao and Li, Zhaohai and Li, Zhixun and Gu, Tiancheng and Wei, Xilin and Zang, Yuhang and Li, Hongsheng and Lin, Dahua and Zhang, Guofeng and Bai, Shuai and Yang, Zhibo and Liu, Dayiheng},
  title = {{BlenderWorld}: Benchmarking Visual Agents for {3D} Creation from Real-World References},
  journal = {arXiv preprint arXiv:2610.23986},
  year = {2026},
  url = {https://arxiv.org/abs/2610.23986}
}

@inproceedings{khan2024text2cad,
  author = {Khan, Mohammad Sadil and Sinha, Sankalp and Sheikh, Talha Uddin and Stricker, Didier and Ali, Sk Aziz and Afzal, Muhammad Zeshan},
  title = {{Text2CAD}: Generating Sequential {CAD} Designs from Beginner-to-Expert Level Text Prompts},
  booktitle = NIPS,
  volume = {37},
  pages = {7552--7579},
  year = {2024},
  url = {https://proceedings.neurips.cc/paper_files/paper/2024/hash/0e5b96f97c1813bb75f6c28532c2ecc7-Abstract.html}
}

@inproceedings{rukhovich2025cadrecode,
  author = {Rukhovich, Danila and Dupont, Elona and Mallis, Dimitrios and Cherenkova, Kseniya and Kacem, Anis and Aouada, Djamila},
  title = {{CAD-Recode}: Reverse Engineering {CAD} Code from Point Clouds},
  booktitle = ICCV,
  pages = {9801--9811},
  year = {2025},
  url = {https://openaccess.thecvf.com/content/ICCV2025/html/Rukhovich_CAD-Recode_Reverse_Engineering_CAD_Code_from_Point_Clouds_ICCV_2025_paper.html}
}

@inproceedings{wang2025cadfusion,
  author = {Wang, Ruiyu and Yuan, Yu and Sun, Shizhao and Bian, Jiang},
  title = {Text-to-{CAD} Generation Through Infusing Visual Feedback in Large Language Models},
  booktitle = {ICML},
  pages = {65326--65345},
  year = {2025},
  url = {https://proceedings.mlr.press/v267/wang25eg.html}
}

@inproceedings{mallis2025cadassistant,
  author = {Mallis, Dimitrios and Karadeniz, Ahmet Serdar and Cavada, Sebastian and Rukhovich, Danila and Foteinopoulou, Niki and Cherenkova, Kseniya and Kacem, Anis and Aouada, Djamila},
  title = {{CAD-Assistant}: Tool-Augmented {VLLMs} as Generic {CAD} Task Solvers},
  booktitle = ICCV,
  year = {2025},
  url = {https://arxiv.org/abs/2412.13810}
}

@article{hu2026itercad,
  author = {Hu, Tao and Ai, Jiaxin and Wen, Licheng and Li, Xueheng and Zou, Shu and Li, Siqi and Deng, Nianchen and Cai, Xinyu and Zhou, Hongbin and Cai, Pinlong and Fu, Daocheng and Yang, Yu and Zhang, Hairong and Shi, Botian and Yang, Xuemeng},
  title = {{IterCAD}: An Iterative Multimodal Agent for Visually-Grounded {CAD} Generation and Editing},
  journal = {arXiv preprint arXiv:2606.13368},
  year = {2026},
  url = {https://arxiv.org/abs/2606.13368}
}

@article{xu2026artisancad,
  author = {Xu, Yunhan and Wu, Qifeng and Li, Xunjin and Bin, Yuanwei and Yao, Qingsong and Gu, Jianghang and Wang, Guan and Lv, Weihao and Yang, Huiyu and Luo, Wenfa and Xiang, Jiao and Chen, Yuntian and Chen, Shiyi},
  title = {{ArtisanCAD}: An Industrial-Level {CAD} Agent with Expert-Grounded Knowledge Distillation},
  journal = {arXiv preprint arXiv:2607.05750},
  year = {2026},
  url = {https://arxiv.org/abs/2607.05750}
}

@article{yang2026p3dbench,
  author = {Yang, Yikang and Hu, Zhanpeng and Lin, Youtian and Zhou, Mengqi and Xu, Jingxi and Zhang, Feihu and Liu, Jiaheng and Yao, Yao},
  title = {{P3D-Bench}: Benchmarking {MLLMs} for Parametric {3D} Generation and Structural Reasoning},
  journal = {arXiv preprint arXiv:2606.11152},
  year = {2026},
  url = {https://arxiv.org/abs/2606.11152}
}

@inproceedings{blanz1999morphable,
  author    = {Blanz, Volker and Vetter, Thomas},
  title     = {A morphable model for the synthesis of {3D} faces},
  booktitle = {SIGGRAPH},
  pages     = {187--194},
  year      = {1999}
}

@article{li2017learning,
  author    = {Li, Tianye and Bolkart, Timo and Black, Michael J. and Li, Hao and Romero, Javier},
  title     = {Learning a model of facial shape and expression from {4D} scans},
  journal   = TOG,
  volume    = {36},
  number    = {6},
  pages     = {194:1--194:17},
  year      = {2017}
}

@inproceedings{kit2020ict,
  author    = {Li, Ruilong and Bladin, Karl and Zhao, Yajie and Chinara, Chinmay and Ingraham, Owen and Xiang, Pengda and Ren, Xinglei and Prasad, Pratusha and Kishore, Bipin and Xing, Jun and Li, Hao},
  title     = {Learning formation of physically-based face attributes},
  booktitle = CVPR,
  pages     = {3410--3419},
  year      = {2020}
}

@inproceedings{yang2020facescape,
  author    = {Yang, Haotian and Zhu, Hao and Wang, Yanru and Huang, Mingkai and Shen, Qiu and Yang, Ruigang and Cao, Xun},
  title     = {{FaceScape}: A large-scale high quality {3D} face dataset and detailed riggable {3D} face prediction},
  booktitle = CVPR,
  pages     = {601--610},
  year      = {2020}
}

@article{feng2021learning,
  author    = {Feng, Yao and Feng, Haiwen and Black, Michael J. and Bolkart, Timo},
  title     = {Learning an animatable detailed {3D} face model from in-the-wild images},
  journal   = TOG,
  volume    = {40},
  number    = {4},
  pages     = {88:1--88:13},
  year      = {2021}
}

@inproceedings{sanyal2019learning,
  author    = {Sanyal, Soubhik and Bolkart, Timo and Feng, Haiwen and Black, Michael J.},
  title     = {Learning to regress {3D} face shape and expression from an image without {3D} supervision},
  booktitle = CVPR,
  pages     = {7763--7772},
  year      = {2019}
}

@inproceedings{danvevcek2022emoca,
  author    = {Dan{\v{e}}{\v{c}}ek, Radek and Black, Michael J. and Bolkart, Timo},
  title     = {{EMOCA}: Emotion driven monocular face capture and animation},
  booktitle = CVPR,
  pages     = {20311--20322},
  year      = {2022}
}

@inproceedings{zielonka2022towards,
  author    = {Zielonka, Wojciech and Bolkart, Timo and Thies, Justus},
  title     = {Towards metrical reconstruction of human faces},
  booktitle = ECCV,
  pages     = {250--269},
  year      = {2022}
}

@inproceedings{retsinas2024smirk,
  author    = {Retsinas, Georgios and Filntisis, Panagiotis P. and Danecek, Radek and Abrevaya, Victoria F. and Roussos, Anastasios and Bolkart, Timo and Maragos, Petros},
  title     = {{3D} facial expressions through analysis-by-neural-synthesis},
  booktitle = CVPR,
  pages     = {2490--2501},
  year      = {2024}
}

@techreport{hunyuan3d22025,
  author      = {{Tencent Hunyuan3D Team}},
  title       = {{Hunyuan3D 2.0}: Scaling diffusion models for high resolution textured {3D} assets generation},
  institution = {Tencent},
  number      = {arXiv:2501.12202},
  year        = {2025},
  url         = {https://arxiv.org/abs/2501.12202}
}

@inproceedings{xiang2024structured,
  author    = {Xiang, Jianfeng and Lv, Zelong and Xu, Sicheng and Deng, Yu and Wang, Ruicheng and Zhang, Bowen and Chen, Dong and Tong, Xin and Yang, Jiaolong},
  title     = {Structured {3D} latents for scalable and versatile {3D} generation},
  booktitle = CVPR,
  year      = {2025}
}

@inproceedings{hong2023lrm,
  author    = {Hong, Yicong and Zhang, Kai and Gu, Jiuxiang and Bi, Sai and Zhou, Yang and Liu, Difan and Liu, Feng and Sunkavalli, Kalyan and Bui, Trung and Tan, Hao},
  title     = {{LRM}: Large reconstruction model for single image to {3D}},
  booktitle = ICLR,
  year      = {2024}
}

@inproceedings{jimenez2024swebench,
  author    = {Jimenez, Carlos E. and Yang, John and Wettig, Alexander and Yao, Shunyu and Pei, Kexin and Press, Ofir and Narasimhan, Karthik},
  title     = {{SWE-bench}: Can language models resolve real-world {GitHub} issues?},
  booktitle = ICLR,
  year      = {2024}
}

@misc{terminalbench2025,
  author    = {{Terminal-Bench Team}},
  title     = {{Terminal-Bench}: A benchmark for multimodal and terminal coding agents via {Harbor}},
  year      = {2025},
  note      = {arXiv preprint}
}

@article{du2024blenderllm,
  author = {Du, Yuhao and Chen, Shunian and Zan, Wenbo and Li, Peizhao and Wang, Mingxuan and Song, Dingjie and Li, Bo and Hu, Yan and Wang, Benyou},
  title = {{BlenderLLM}: Training large language models for computer-aided design with self-improvement},
  journal = {arXiv preprint arXiv:2412.14203},
  year = {2024}
}

@inproceedings{gu2024bladergym,
  author    = {Gu, Yunqi and Huang, Ian and Je, Jihyeon and Yang, Guandao and Guibas, Leonidas},
  title     = {{BlenderGym}: Benchmarking foundational model systems for graphics editing},
  booktitle = CVPR,
  year      = {2025}
}

@inproceedings{hu2024scenecraft,
  author    = {Hu, Ziniu and Iscen, Ahmet and Jain, Aashi and Kipf, Thomas and Yue, Yisong and Ross, David A. and Schmid, Cordelia and Fathi, Alireza},
  title     = {{SceneCraft}: An {LLM} agent for synthesizing {3D} scenes as {Blender} code},
  booktitle = {ICML},
  year      = {2024}
}

@article{sun20233dgpt,
  author    = {Sun, Chunyi and Han, Junlin and Deng, Weijian and Wang, Xinlong and Qin, Zishan and Gould, Stephen},
  title     = {{3D-GPT}: Procedural {3D} modeling with large language models},
  journal   = {arXiv preprint arXiv:2310.12945},
  year      = {2023}
}

@article{knapitsch2017tanks,
  author    = {Knapitsch, Arno and Park, Jaesik and Zhou, Qian-Yi and Koltun, Vladlen},
  title     = {Tanks and temples: Benchmarking large-scale scene reconstruction},
  journal   = TOG,
  volume    = {36},
  number    = {4},
  pages     = {78:1--78:13},
  year      = {2017}
}

@inproceedings{tatarchenko2019single,
  author    = {Tatarchenko, Maxim and Richter, Stephan R. and Ranftl, Ren{\'e} and Li, Zhuwen and Koltun, Vladlen and Brox, Thomas},
  title     = {What do single-view {3D} reconstruction networks learn?},
  booktitle = CVPR,
  pages     = {3405--3414},
  year      = {2019}
}

@misc{openai2026codexcli,
  author = {{OpenAI}},
  title = {{Codex CLI}},
  howpublished = {Official software repository. \url{https://github.com/openai/codex}},
  url = {https://github.com/openai/codex},
  year = {2026},
  note = {Version 0.162.1 used in experiments; accessed 2026-11-05}
}

@misc{anomaly2026opencode,
  author = {{Anomaly}},
  title = {{OpenCode}},
  howpublished = {Official software repository. \url{https://github.com/anomalyco/opencode}},
  url = {https://github.com/anomalyco/opencode},
  year = {2026},
  note = {Version 1.18.35 used in experiments; accessed 2026-11-05}
}

@misc{nous2026hermesagent,
  author = {{Nous Research}},
  title = {{Hermes Agent}},
  howpublished = {Official software repository. \url{https://github.com/NousResearch/hermes-agent}},
  url = {https://github.com/NousResearch/hermes-agent},
  year = {2026},
  note = {Version 0.21.6 used in experiments; accessed 2026-11-05}
}
}

\onecolumn
\appendix
\supplementlayout
\linespread{1.04}\selectfont
\setcounter{table}{0}
\setcounter{figure}{0}
\renewcommand{\thetable}{S\arabic{table}}
\renewcommand{\thefigure}{S\arabic{figure}}
\renewcommand{\theHtable}{supp.\arabic{table}}
\renewcommand{\theHfigure}{supp.\arabic{figure}}

\pdfbookmark[0]{Supplementary Material}{supplement}
\begin{center}
    \Large\bfseries Supplementary Material
\end{center}
\section{Supplementary Material and Reading Guide}
This material provides the complete metrics, cross-domain results, initialization and harness comparisons, and session diagnostics omitted from the main text for space. Section~\ref{sec:s_diagnosis} reports the plan replay and single-step handoff experiments. Tables and figures are numbered separately in the main text and supplementary material.

\section{Complete Main-Experiment Metrics}
\begin{figure}[H]
\centering
\includegraphics[width=\textwidth]{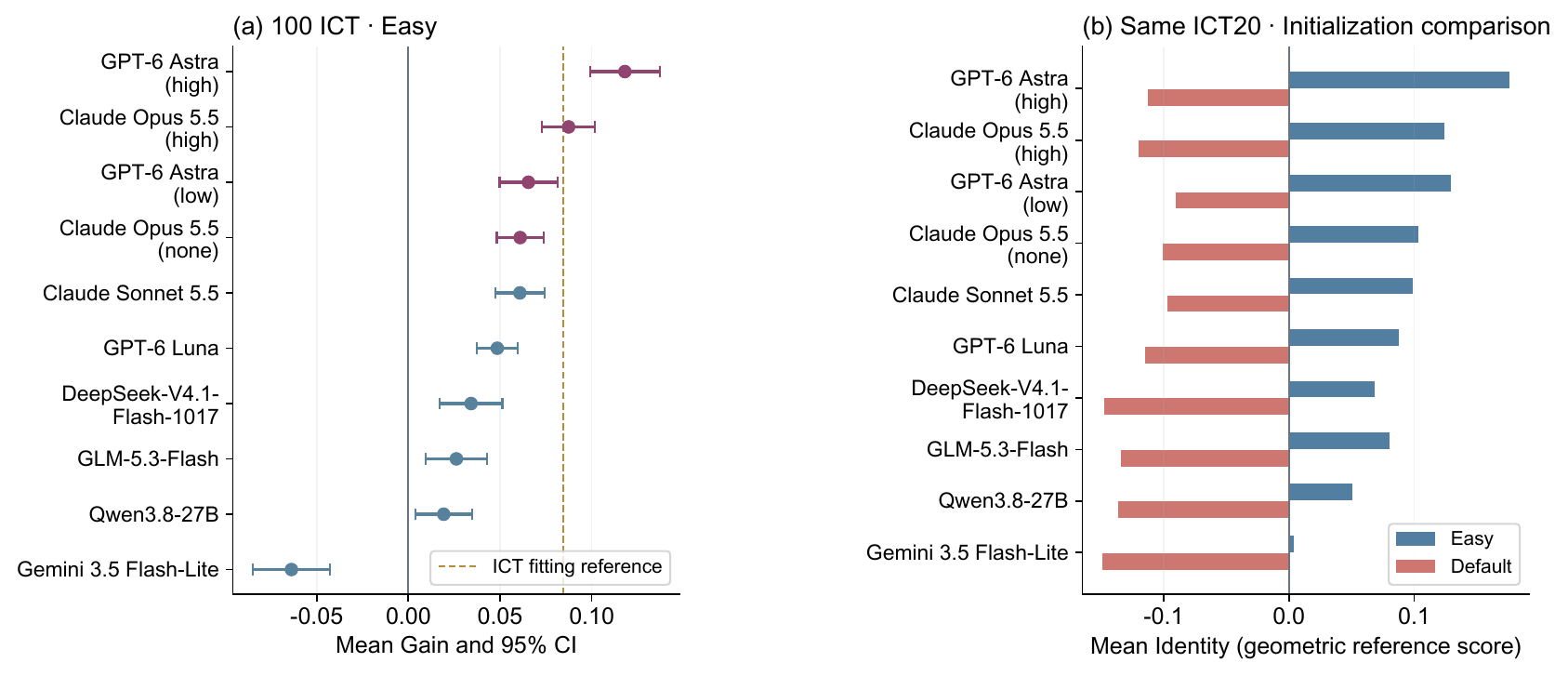}
\caption{Improvements from mean-face editing and initialization differences. (a) Mean Gain and 95\% CI on the 100 ICT main-set cases; the dashed line denotes the dedicated ICT frontal-fitting reference. (b) Initialization comparison on the same ICT20; positive Identity means only that the score exceeds the in-domain random-individual reference. The settings jointly change shape, topology, and placement priors.}
\label{fig:evidence}
\end{figure}

Table~\ref{tab:s_full} reports whole-head metrics; Table~\ref{tab:s_regions} reports facial-region results for the same predictions. Both list Easy on ICT100 and Default on ICT20; paired initialization comparisons on the same ICT20 appear in Table~\ref{tab:s_start}. Figure~\ref{fig:evidence} visualizes main-set Gain and the paired initialization comparison. Reference methods are excluded from agent rankings, and their Gain is computed relative to the mean face for the same sample. Gain is not applicable to empty-scene agents.

\vspace{-4pt}
\begin{table}[H]
\centering
\caption{Complete whole-head metrics: Easy (ICT100) and Default (ICT20, empty scene). The relative score is Identity from the main text; Above init. counts samples. References are aggregated over the corresponding cases, with Gain and Above init. relative to the mean face; these two quantities are not applicable to empty-scene agents.}
\label{tab:s_full}
{\linespread{1.0}\selectfont\small\setlength{\tabcolsep}{2.2pt}\setlength{\aboverulesep}{0.30ex}\setlength{\belowrulesep}{0.38ex}\renewcommand{\arraystretch}{0.93}
\begin{tabular*}{\textwidth}{@{\extracolsep{\fill}}lccccccc@{}}
\toprule
Method & \fstrict & \floose & Held-out IoU & Identity & Identity$>0$ & \gain [95\% CI] & \makecell[cb]{Above\\init.} \\
\midrule
\suppgrp{8}{Easy (ICT100)} \\
\supptabind GPT-6 Astra (high)$^*$ & \textbf{0.321} & \textbf{0.797} & 0.898 & $\mathbf{+0.174}$ & 97/100 & $\mathbf{+0.118}$\ci{[+0.099, +0.137]} & 87 \\
\supptabind Claude Opus 5.5 (high) & 0.297 & 0.773 & \textbf{0.900} & $+0.145$ & 97/100 & $+0.087$\ci{[+0.073, +0.102]} & \textbf{90} \\
\supptabind GPT-6 Astra (low) & 0.281 & 0.764 & 0.892 & $+0.126$ & \textbf{98/100} & $+0.065$\ci{[+0.050, +0.081]} & 87 \\
\supptabind Claude Opus 5.5 (none) & 0.277 & 0.757 & 0.889 & $+0.120$ & 95/100 & $+0.061$\ci{[+0.048, +0.074]} & \textbf{90} \\
\supptabind Claude Sonnet 5.5 & 0.277 & 0.758 & 0.887 & $+0.121$ & 97/100 & $+0.061$\ci{[+0.048, +0.074]} & 84 \\
\supptabind GPT-6 Luna & 0.268 & 0.750 & 0.885 & $+0.109$ & \textbf{98/100} & $+0.049$\ci{[+0.037, +0.060]} & 81 \\
\supptabind DeepSeek-V4.1-Flash-1017 & 0.258 & 0.726 & 0.891 & $+0.097$ & 91/100 & $+0.034$\ci{[+0.017, +0.051]} & 67 \\
\supptabind GLM-5.3-Flash & 0.251 & 0.718 & 0.867 & $+0.088$ & 87/100 & $+0.026$\ci{[+0.009, +0.043]} & 54 \\
\supptabind Qwen3.8-27B & 0.245 & 0.711 & 0.879 & $+0.082$ & 84/100 & $+0.019$\ci{[+0.004, +0.035]} & 56 \\
\supptabind Gemini 3.5 Flash-Lite$^*$ & 0.182 & 0.587 & 0.791 & $+0.005$ & 60/100 & $-0.064$\ci{[-0.085, -0.043]} & 13 \\
\supptabind DECA & 0.219 & 0.651 & 0.876 & $+0.050$ & 80/100 & $-0.016$\ci{[-0.028, -0.003]} & 36 \\
\supptabind Hunyuan3D-2 & 0.190 & 0.551 & 0.817 & $+0.014$ & 61/100 & $-0.057$\ci{[-0.076, -0.038]} & 27 \\
\supptabind \textit{Mean-face initialization ($\initf$)} & \textit{0.229} & \textit{0.702} & \textit{0.869} & $\mathit{+0.062}$ & \textit{83/100} & \gainonly{$\mathit{\phantom{+}0.000}$} & --- \\
\supptabind \textit{ICT 3DMM frontal fitting} & \textit{0.297} & \textit{0.760} & \textit{0.909} & $\mathit{+0.144}$ & \textit{94/100} & $\mathit{+0.084}$\ci{[+0.063, +0.106]} & \textit{78} \\
\supptabind \textit{10-configuration per-sample best envelope} & \textit{0.342} & \textit{0.812} & --- & $\mathit{+0.198}$ & \textit{99/100} & \gainonly{$\mathit{+0.145}$} & \textit{99} \\
\midrule
\suppgrp{8}{Default (ICT20, empty scene)} \\
\supptabind GPT-6 Astra (high)$^*$ & 0.084 & 0.312 & 0.683 & $-0.113$ & 1/20 & --- & --- \\
\supptabind Claude Opus 5.5 (high) & 0.078 & 0.295 & 0.672 & $-0.120$ & 0/20 & --- & --- \\
\supptabind GPT-6 Astra (low) & \textbf{0.103} & \textbf{0.374} & 0.658 & $\mathbf{-0.090}$ & 0/20 & --- & --- \\
\supptabind Claude Opus 5.5 (none) & 0.094 & 0.346 & 0.672 & $-0.101$ & \textbf{2/20} & --- & --- \\
\supptabind Claude Sonnet 5.5 & 0.097 & 0.359 & \textbf{0.684} & $-0.097$ & 0/20 & --- & --- \\
\supptabind GPT-6 Luna & 0.082 & 0.314 & 0.629 & $-0.115$ & 0/20 & --- & --- \\
\supptabind DeepSeek-V4.1-Flash-1017 & 0.055 & 0.206 & 0.520 & $-0.147$ & 0/20 & --- & --- \\
\supptabind GLM-5.3-Flash & 0.067 & 0.252 & 0.552 & $-0.134$ & 0/20 & --- & --- \\
\supptabind Qwen3.8-27B & 0.065 & 0.274 & 0.640 & $-0.136$ & 0/20 & --- & --- \\
\supptabind Gemini 3.5 Flash-Lite$^*$ & 0.054 & 0.212 & 0.619 & $-0.149$ & 0/20 & --- & --- \\
% BEGIN MATCHED ICT20 REFERENCES
\midrule
\suppgrp{8}{References on the same ICT20} \\
\supptabind DECA & 0.206 & 0.619 & 0.867 & $+0.036$ & 14/20 & $-0.028$\ci{[-0.052, -0.003]} & 5 \\
\supptabind Hunyuan3D-2 & 0.174 & 0.518 & 0.808 & $-0.004$ & 10/20 & $-0.072$\ci{[-0.113, -0.031]} & 3 \\
\supptabind \textit{Mean-face initialization} & \textit{0.226} & \textit{0.688} & \textit{0.865} & $\mathit{+0.061}$ & \textit{18/20} & \gainonly{$\mathit{\phantom{+}0.000}$} & --- \\
\supptabind \textit{ICT 3DMM frontal fitting} & \textit{0.292} & \textit{0.738} & \textit{0.905} & $\mathit{+0.140}$ & \textit{19/20} & $\mathit{+0.082}$\ci{[+0.030, +0.135]} & \textit{14} \\
% END MATCHED ICT20 REFERENCES
\bottomrule
\end{tabular*}}
\end{table}

\clearpage
\begin{table}[H]
\centering
\caption{Facial and facial-region metrics. Easy uses ICT100 and Default uses ICT20; regional columns report F@0.5\%, and normal error is in degrees.}
\label{tab:s_regions}
{\linespread{1.0}\selectfont\small\setlength{\tabcolsep}{3pt}\renewcommand{\arraystretch}{1.12}
\begin{tabular*}{\textwidth}{@{\extracolsep{\fill}}lcccccccc@{}}
\toprule
Method & Facial F & Facial Gain & Brow/eyes & Nose & Mouth & Jaw & Ears & Normal \\
\midrule
\suppgrp{9}{Easy (ICT100)} \\[1pt]
\supptabind GPT-6 Astra (high)$^*$ & \textbf{0.411} & $\mathbf{+0.200}$ & \textbf{0.365} & \textbf{0.348} & \textbf{0.443} & \textbf{0.379} & \textbf{0.478} & \textbf{28.3} \\
\supptabind Claude Opus 5.5 (high) & 0.351 & $+0.115$ & 0.299 & 0.274 & 0.324 & 0.375 & 0.409 & 29.7 \\
\supptabind GPT-6 Astra (low) & 0.358 & $+0.126$ & 0.341 & 0.286 & 0.368 & 0.366 & 0.382 & 29.1 \\
\supptabind Claude Opus 5.5 (none) & 0.316 & $+0.070$ & 0.270 & 0.249 & 0.297 & 0.364 & 0.357 & 30.4 \\
\supptabind Claude Sonnet 5.5 & 0.317 & $+0.070$ & 0.276 & 0.261 & 0.301 & 0.353 & 0.348 & 30.4 \\
\supptabind GPT-6 Luna & 0.309 & $+0.060$ & 0.292 & 0.275 & 0.316 & 0.349 & 0.320 & 30.4 \\
\supptabind DeepSeek-V4.1-Flash-1017 & 0.277 & $+0.011$ & 0.266 & 0.245 & 0.275 & 0.290 & 0.299 & 33.6 \\
\supptabind GLM-5.3-Flash & 0.276 & $+0.013$ & 0.261 & 0.249 & 0.278 & 0.355 & 0.262 & 32.0 \\
\supptabind Qwen3.8-27B & 0.259 & $-0.011$ & 0.246 & 0.255 & 0.269 & 0.323 & 0.250 & 34.4 \\
\supptabind Gemini 3.5 Flash-Lite$^*$ & 0.208 & $-0.080$ & 0.223 & 0.223 & 0.258 & 0.308 & 0.140 & 34.4 \\
\supptabind DECA & 0.256 & $-0.012$ & 0.172 & 0.269 & 0.256 & 0.323 & 0.284 & 30.8 \\
\supptabind Hunyuan3D-2 & 0.177 & $-0.129$ & 0.118 & 0.092 & 0.149 & 0.161 & 0.249 & 41.0 \\
\supptabind \textit{Mean-face initialization ($\initf$)} & \textit{0.262} & $\mathit{\phantom{+}0.000}$ & \textit{0.270} & \textit{0.270} & \textit{0.291} & \textit{0.360} & \textit{0.207} & \textit{30.7} \\
\supptabind \textit{ICT 3DMM frontal fitting} & \textit{0.303} & $\mathit{+0.045}$ & \textit{0.233} & \textit{0.227} & \textit{0.283} & \textit{0.326} & \textit{0.372} & \textit{30.6} \\
\midrule
\suppgrp{9}{Default (ICT20, empty scene)} \\[1pt]
\supptabind GPT-6 Astra (high)$^*$ & 0.132 & --- & 0.081 & 0.018 & 0.029 & 0.029 & \textbf{0.259} & --- \\
\supptabind Claude Opus 5.5 (high) & 0.089 & --- & 0.002 & 0.024 & 0.033 & 0.003 & 0.212 & --- \\
\supptabind GPT-6 Astra (low) & \textbf{0.139} & --- & \textbf{0.120} & 0.038 & 0.028 & 0.043 & 0.224 & --- \\
\supptabind Claude Opus 5.5 (none) & 0.116 & --- & 0.002 & 0.013 & 0.051 & \textbf{0.055} & 0.257 & --- \\
\supptabind Claude Sonnet 5.5 & 0.103 & --- & 0.003 & 0.014 & 0.022 & \textbf{0.055} & 0.238 & --- \\
\supptabind GPT-6 Luna & 0.113 & --- & 0.080 & 0.060 & \textbf{0.058} & 0.049 & 0.181 & --- \\
\supptabind DeepSeek-V4.1-Flash-1017 & 0.073 & --- & 0.046 & 0.016 & 0.050 & 0.050 & 0.107 & --- \\
\supptabind GLM-5.3-Flash & 0.086 & --- & 0.024 & 0.052 & 0.035 & 0.038 & 0.152 & --- \\
\supptabind Qwen3.8-27B & 0.095 & --- & 0.047 & \textbf{0.071} & 0.039 & 0.023 & 0.159 & --- \\
\supptabind Gemini 3.5 Flash-Lite$^*$ & 0.027 & --- & 0.008 & 0.013 & 0.028 & 0.006 & 0.046 & --- \\
% BEGIN MATCHED ICT20 REFERENCES
\midrule
\suppgrp{9}{References on the same ICT20} \\[1pt]
\supptabind DECA & 0.274 & $-0.009$ & 0.183 & 0.311 & 0.228 & 0.303 & 0.330 & 30.1 \\
\supptabind Hunyuan3D-2 & 0.163 & $-0.174$ & 0.114 & 0.100 & 0.068 & 0.128 & 0.230 & 40.4 \\
\supptabind \textit{Mean-face initialization} & \textit{0.279} & $\mathit{+0.000}$ & \textit{0.276} & \textit{0.279} & \textit{0.282} & \textit{0.422} & \textit{0.229} & \textit{30.0} \\
\supptabind \textit{ICT 3DMM frontal fitting} & \textit{0.294} & $\mathit{+0.011}$ & \textit{0.233} & \textit{0.210} & \textit{0.253} & \textit{0.308} & \textit{0.356} & \textit{31.0} \\
% END MATCHED ICT20 REFERENCES
\bottomrule
\end{tabular*}}
\end{table}

\paragraph{Reasoning effort and statistical scope.} Configurations including upstream waiting time are marked $^*$. GPT-6 Luna uses xhigh; Claude Sonnet 5.5 and Gemini 3.5 Flash-Lite use high. GPT-6 Astra uses low/high, and Claude Opus 5.5 uses none/high. The three self-hosted models use their respective recommended high-reasoning settings. Confidence intervals in the main text are $95\%$ t intervals for the mean per-sample Gain; paired model comparisons use the Wilcoxon signed-rank test with Holm correction within each comparison family. Intervals over the 100 tasks reflect between-task variation, not randomness across repeated runs of a single task.

\clearpage

\vspace{6pt}
\section{Cross-Domain Results and Reference Scores}
Agents use the ICT mean face as initialization in every domain. The leave-one-out in-domain mean face achieves F of 0.518 on FaceScape, above the strongest listed prediction at 0.438; large Gain therefore cannot be interpreted as recovery of individual identity. The model-family match between FLAME and DECA, and head-shape differences in MPFB, may also affect cross-domain rankings.

\begin{table}[H]
\centering
\caption{Representative methods on the three cross-domain subsets ($n=10$ each).}
\label{tab:s_xdomain}
{\linespread{1.0}\selectfont\small\setlength{\tabcolsep}{3pt}\renewcommand{\arraystretch}{1.12}
\begin{tabular*}{\textwidth}{@{\extracolsep{\fill}}llcccccc@{}}
\toprule
Source & Method & \fstrict & \ident & Identity$>0$ & \gain & Facial F & Facial Gain \\
\midrule
\multirow{6}{*}{FaceScape}
 & Claude Opus 5.5 (high) & \textbf{0.438} & $\mathbf{+0.016}$ & \textbf{5/10} & $\mathbf{+0.302}$ & \textbf{0.457} & $\mathbf{+0.317}$ \\
 & GPT-6 Astra (high)$^*$ & 0.428 & $-0.003$ & \textbf{5/10} & $+0.293$ & 0.413 & $+0.258$ \\
 & Claude Sonnet 5.5 & 0.384 & $-0.073$ & 3/10 & $+0.235$ & 0.407 & $+0.243$ \\
 & DeepSeek-V4.1-Flash-1017 & 0.347 & $-0.139$ & 3/10 & $+0.181$ & 0.355 & $+0.176$ \\
 & DECA & 0.270 & $-0.273$ & 0/10 & $+0.094$ & 0.255 & $+0.054$ \\
 & Hunyuan3D-2 & 0.263 & $-0.286$ & 0/10 & $+0.080$ & 0.230 & $+0.020$ \\
\midrule
\multirow{4}{*}{FLAME}
 & GPT-6 Astra (high)$^*$ & \textbf{0.288} & $\mathbf{+0.127}$ & \textbf{10/10} & $\mathbf{+0.089}$ & \textbf{0.347} & $\mathbf{+0.132}$ \\
 & DECA & 0.268 & $+0.102$ & 9/10 & $+0.062$ & 0.262 & $+0.018$ \\
 & Claude Opus 5.5 (none) & 0.252 & $+0.084$ & 9/10 & $+0.045$ & 0.245 & $-0.005$ \\
 & Hunyuan3D-2 & 0.209 & $+0.030$ & 7/10 & $-0.013$ & 0.170 & $-0.118$ \\
\midrule
\multirow{5}{*}{MPFB}
 & Hunyuan3D-2 & \textbf{0.286} & $\mathbf{+0.147}$ & \textbf{10/10} & $\mathbf{+0.134}$ & 0.229 & $-0.004$ \\
 & GPT-6 Astra (high)$^*$ & 0.236 & $+0.087$ & 9/10 & $+0.074$ & 0.289 & $+0.083$ \\
 & Claude Opus 5.5 (high) & 0.234 & $+0.085$ & 8/10 & $+0.075$ & \textbf{0.293} & $\mathbf{+0.120}$ \\
 & Qwen3.8-27B & 0.228 & $+0.078$ & 9/10 & $+0.066$ & 0.248 & $+0.054$ \\
 & DECA & 0.219 & $+0.068$ & 8/10 & $+0.061$ & 0.218 & $+0.039$ \\
\bottomrule
\end{tabular*}}
\end{table}

\clearpage
\section{Complete Initialization and Harness Comparisons}
Initializations are compared on the same ICT20. Easy and Default jointly change geometric priors, topology, and canonical placement, so their difference cannot be attributed solely to visual perception.

\begin{table}[H]
\centering
\caption{Initialization comparison on ICT20. The relative score is Identity calibrated against in-domain random individuals; the difference is Easy minus Default.}
\label{tab:s_start}
{\linespread{1.0}\selectfont\small\setlength{\tabcolsep}{3pt}\renewcommand{\arraystretch}{1.07}
\begin{tabular*}{\textwidth}{@{\extracolsep{\fill}}lcccccc@{}}
\toprule
Configuration & \makecell[cb]{Default\\F} & \makecell[cb]{Default\\facial F} & \makecell[cb]{Default\\Identity} & Identity$>0$ & \makecell[cb]{Easy\\Identity} & Difference \\
\midrule
GPT-6 Astra (high)$^*$ & 0.084 & 0.132 & $-0.113$ & 1/20 & $\mathbf{+0.176}$ & $\mathbf{+0.288}$ \\
Claude Opus 5.5 (high) & 0.078 & 0.089 & $-0.120$ & 0/20 & $+0.124$ & $+0.244$ \\
GPT-6 Astra (low) & \textbf{0.103} & \textbf{0.139} & $\mathbf{-0.090}$ & 0/20 & $+0.130$ & $+0.220$ \\
Claude Opus 5.5 (none) & 0.094 & 0.116 & $-0.101$ & \textbf{2/20} & $+0.103$ & $+0.204$ \\
Claude Sonnet 5.5 & 0.097 & 0.103 & $-0.097$ & 0/20 & $+0.099$ & $+0.196$ \\
GPT-6 Luna & 0.082 & 0.113 & $-0.115$ & 0/20 & $+0.088$ & $+0.202$ \\
DeepSeek-V4.1-Flash-1017 & 0.055 & 0.073 & $-0.147$ & 0/20 & $+0.069$ & $+0.216$ \\
GLM-5.3-Flash & 0.067 & 0.086 & $-0.134$ & 0/20 & $+0.081$ & $+0.215$ \\
Qwen3.8-27B & 0.065 & 0.095 & $-0.136$ & 0/20 & $+0.051$ & $+0.187$ \\
Gemini 3.5 Flash-Lite & 0.054 & 0.027 & $-0.149$ & 0/20 & $+0.004$ & $+0.153$ \\
\bottomrule
\end{tabular*}}
\end{table}

\begin{table}[H]
\centering
\caption{Four-harness comparison on ICT20. Each cell reports whole-head Gain / facial Gain; the final column gives the configuration-specific Friedman test.}
\label{tab:s_harness}
{\linespread{1.0}\selectfont\small\setlength{\tabcolsep}{3pt}\renewcommand{\arraystretch}{1.07}
\begin{tabular*}{\textwidth}{@{\extracolsep{\fill}}lccccc@{}}
\toprule
Configuration & terminus-2-vision & Codex & OpenCode & Hermes & $p$ \\
\midrule
GPT-6 Astra (high)$^*$ & \hgain{$+0.121$}{$+0.225$} & \hgain{$\mathbf{+0.122}$}{$\mathbf{+0.241}$} & \hgain{$+0.081$}{$+0.131$} & \hgain{$+0.116$}{$+0.186$} & \textbf{0.017} \\
Claude Opus 5.5 (high) & \hgain{$+0.067$}{$+0.131$} & \hgain{$+0.081$}{$+0.120$} & \hgain{$+0.071$}{$+0.068$} & \hgain{$+0.073$}{$+0.095$} & 0.90 \\
GPT-6 Astra (low) & \hgain{$+0.072$}{$+0.150$} & \hgain{$+0.083$}{$+0.138$} & \hgain{$+0.067$}{$+0.131$} & \hgain{$+0.068$}{$+0.126$} & 0.061 \\
Claude Opus 5.5 (none) & \hgain{$+0.045$}{$+0.052$} & \hgaindag{$+0.053$}{$+0.065$} & \hgain{$+0.056$}{$+0.093$} & \hgaindag{$+0.053$}{$+0.082$} & 0.98 \\
Claude Sonnet 5.5 & \hgain{$+0.038$}{$+0.047$} & \hgain{$+0.041$}{$+0.053$} & \hgain{$+0.022$}{$+0.049$} & \hgain{$+0.048$}{$+0.108$} & 0.16 \\
GPT-6 Luna & \hgain{$+0.028$}{$+0.084$} & \hgain{$+0.039$}{$+0.073$} & \hgain{$+0.023$}{$+0.028$} & \hgain{$+0.047$}{$+0.100$} & 0.31 \\
DeepSeek-V4.1-Flash-1017 & \hgain{$+0.006$}{$-0.061$} & \hgain{$+0.018$}{$-0.021$} & \hgain{$+0.043$}{$+0.049$} & \hgain{$+0.039$}{$-0.007$} & 0.65 \\
GLM-5.3-Flash & \hgain{$+0.021$}{$+0.002$} & \hgain{$-0.004$}{$+0.004$} & \hgain{$+0.027$}{$+0.010$} & \hgain{$-0.021$}{$-0.027$} & \textbf{0.007} \\
Qwen3.8-27B & \hgain{$-0.012$}{$-0.041$} & \hgain{$-0.007$}{$-0.051$} & \hgain{$+0.010$}{$-0.034$} & \hgain{$-0.008$}{$-0.035$} & 0.46 \\
Gemini 3.5 Flash-Lite & \hgain{$-0.084$}{$-0.098$} & --- & \hgain{$-0.025$}{$-0.032$} & --- & --- \\
\bottomrule
\end{tabular*}}
\end{table}

% Generated by generate_harness_cost_table.py; do not edit values manually.
\begin{table}[H]
\centering
\caption{Harness-dependent interaction cost on the same ICT20 targets. Each cell reports median wall-clock time (minutes) / median cumulative tokens (millions), including repeated input context. Invalid API combinations are omitted.}
\label{tab:s_harness_cost}
{\linespread{1.0}\selectfont\small\setlength{\tabcolsep}{3pt}\renewcommand{\arraystretch}{1.07}
\begin{tabular*}{\textwidth}{@{\extracolsep{\fill}}lcccc@{}}
\toprule
Model & terminus-2-vision & Codex & opencode & Hermes \\
\midrule
GPT-6 Astra (high) & \hcost{57.7}{4.13} & \hcost{56.6}{4.38} & \hcost{56.9}{6.04} & \hcost{55.2}{6.96} \\
Claude Opus 5.5 (high) & \hcost{17.5}{0.37} & \hcost{18.7}{1.38} & \hcost{21.2}{1.44} & \hcost{25.6}{2.07} \\
GPT-6 Astra (low) & \hcost{4.3}{0.14} & \hcost{16.0}{1.71} & \hcost{6.2}{0.43} & \hcost{10.0}{1.33} \\
Claude Opus 5.5 (none) & \hcost{8.1}{0.16} & \hcost{6.2}{0.48} & \hcost{5.3}{0.43} & \hcost{6.4}{0.58} \\
Claude Sonnet 5.5 & \hcost{4.7}{0.14} & \hcost{2.8}{0.24} & \hcost{1.6}{0.13} & \hcost{2.4}{0.29} \\
GPT-6 Luna & \hcost{7.8}{0.46} & \hcost{23.3}{4.90} & \hcost{13.1}{3.02} & \hcost{9.3}{4.76} \\
DeepSeek-V4.1-Flash-1017 & \hcost{60.0}{5.91} & \hcost{36.7}{6.47} & \hcost{36.0}{6.83} & \hcost{48.3}{10.28} \\
GLM-5.3-Flash & \hcost{60.0}{4.30} & \hcost{4.5}{0.70} & \hcost{18.6}{5.28} & \hcost{17.4}{30.16} \\
Qwen3.8-27B & \hcost{43.8}{3.29} & \hcost{34.4}{5.80} & \hcost{23.8}{9.53} & \hcost{21.8}{8.67} \\
Gemini 3.5 Flash-Lite & \hcost{37.3}{1.29} & --- & \hcost{8.3}{0.75} & --- \\
\bottomrule
\end{tabular*}}
\end{table}

\clearpage
\paragraph{Harness cost.} Table~\ref{tab:s_harness_cost} reports median wall-clock time and cumulative tokens by model and harness on the same ICT20, including waiting and failures in the original runs. Qwen3.8-27B uses 43.8 / 23.8 minutes and 3.29 / 9.53M tokens under \texttt{terminus-2-vision} / OpenCode, respectively, showing that time and token consumption do not move together. Cumulative tokens include repeated input context.

\paragraph{Scope of pooled harness statistics.} Pooling model--sample observations from the nine configurations valid under all four harnesses yields Friedman $p=0.38$, with centered means within $\pm0.002$ for all harnesses. Centering does not change within-group ranks; the same samples are reused across models, so this test is only a descriptive summary, not evidence of harness equivalence. Configuration-specific tests yield $p=0.017$ for GPT-6 Astra (high) and $0.007$ for GLM-5.3-Flash, suggesting that model--harness interactions warrant further study.

\vspace{4pt}
\section{Interaction Cost, Termination, and Per-Sample Correlations}

\begin{table}[H]
\centering
\caption{Interaction cost and within-configuration correlations on the main set (100 ICT). $\rho_{\text{tok}}, \rho_{\text{wall}}$ are within-configuration Spearman correlations of tokens and runtime with Gain. $^{\mathrm{est}}$ denotes estimated OpenRouter API-equivalent cost.}
\label{tab:s_effort}
{\linespread{1.0}\selectfont\small\setlength{\tabcolsep}{2.5pt}\renewcommand{\arraystretch}{1.14}
\begin{tabular*}{\textwidth}{@{\extracolsep{\fill}}lcccccccc@{}}
\toprule
Agent configuration & \makecell[cb]{Same\\score} & \makecell[cb]{Replay\\failures} & \makecell[cb]{Reached\\60m} & \makecell[cb]{Median\\time} & \makecell[cb]{Median\\requests} & \makecell[cb]{Median\\tokens} & \makecell[cb]{Mean \$\\per run} & $\rho_{\text{tok}} / \rho_{\text{wall}}$ \\
\midrule
GPT-6 Astra (high)$^*$ & 2 & 0 & 19 & 57.5\,m & 85.0 & 4.15\,M & \$230.9 & \hcorr{+0.00}{+0.10} \\
Claude Opus 5.5 (high) & 0 & 0 & 0 & 16.7\,m & 20.0 & 0.39\,M & \$9.73 & \hcorr{+0.00}{-0.08} \\
GPT-6 Astra (low) & 0 & 0 & 0 & \textbf{4.1\,m} & \textbf{13.0} & \textbf{0.13\,M} & \$6.09 & \hcorr{-0.09}{-0.05} \\
Claude Opus 5.5 (none) & 0 & 0 & 0 & 8.4\,m & 14.0 & 0.18\,M & \$3.71 & \hcorr{-0.02}{-0.07} \\
Claude Sonnet 5.5 & 1 & 0 & 0 & 6.4\,m & 18.5 & 0.21\,M & \$2.84 & \hcorr{+0.04}{-0.06} \\
GPT-6 Luna & 0 & 0 & 0 & 6.2\,m & 17.0 & 0.33\,M & \textbf{\$0.20} & \hcorr{-0.11}{-0.15} \\
DeepSeek-V4.1-Flash-1017 & 7 & 0 & 44 & 55.0\,m & 124.5 & 6.10\,M & \$2.12$^{\mathrm{est}}$ & \hcorr{+0.01}{-0.03} \\
GLM-5.3-Flash & 27 & 2 & 73 & 60.0\,m & 77.5 & 4.37\,M & \$0.87$^{\mathrm{est}}$ & \hcorr{+0.21}{+0.14} \\
Qwen3.8-27B & 1 & 0 & 20 & 44.0\,m & 76.5 & 3.53\,M & \$0.92$^{\mathrm{est}}$ & \hcorr{-0.12}{-0.10} \\
Gemini 3.5 Flash-Lite$^*$ & 34 & 5 & 15 & 21.0\,m & 50.0 & 1.57\,M & \$1.33 & \hcorr{+0.08}{+0.05} \\
\bottomrule
\end{tabular*}}
\end{table}

\paragraph{Cost accounting.} Tokens are cumulative input and output tokens from request logs, including repeated input context. Except for self-hosted models marked $^{\mathrm{est}}$, cost is total logged cost divided by sample count, affected by API, caching, and routing charges. Time includes model waiting, tool execution, and upstream rate limiting. For self-hosted models, $^{\mathrm{est}}$ estimates API-equivalent cost by pricing cumulative input and output tokens separately for the 100 main-set runs and averaging per run. We use base prices from the \href{https://openrouter.ai/api/v1/models}{OpenRouter models API} on 2026-11-06 (USD per million input / output tokens): 0.15 / 1.875 for Qwen3.8-27B, 0.15 / 0.50 for GLM-5.3-Flash, and 0.30 / 1.20 for DeepSeek-V4.1-Flash-1017. Cached tokens are zero in all three groups; estimates exclude caching, time-based, and routing discounts and do not represent actual GPU self-hosting costs.

\begin{figure}[p]
\centering
\includegraphics[width=0.90\textwidth]{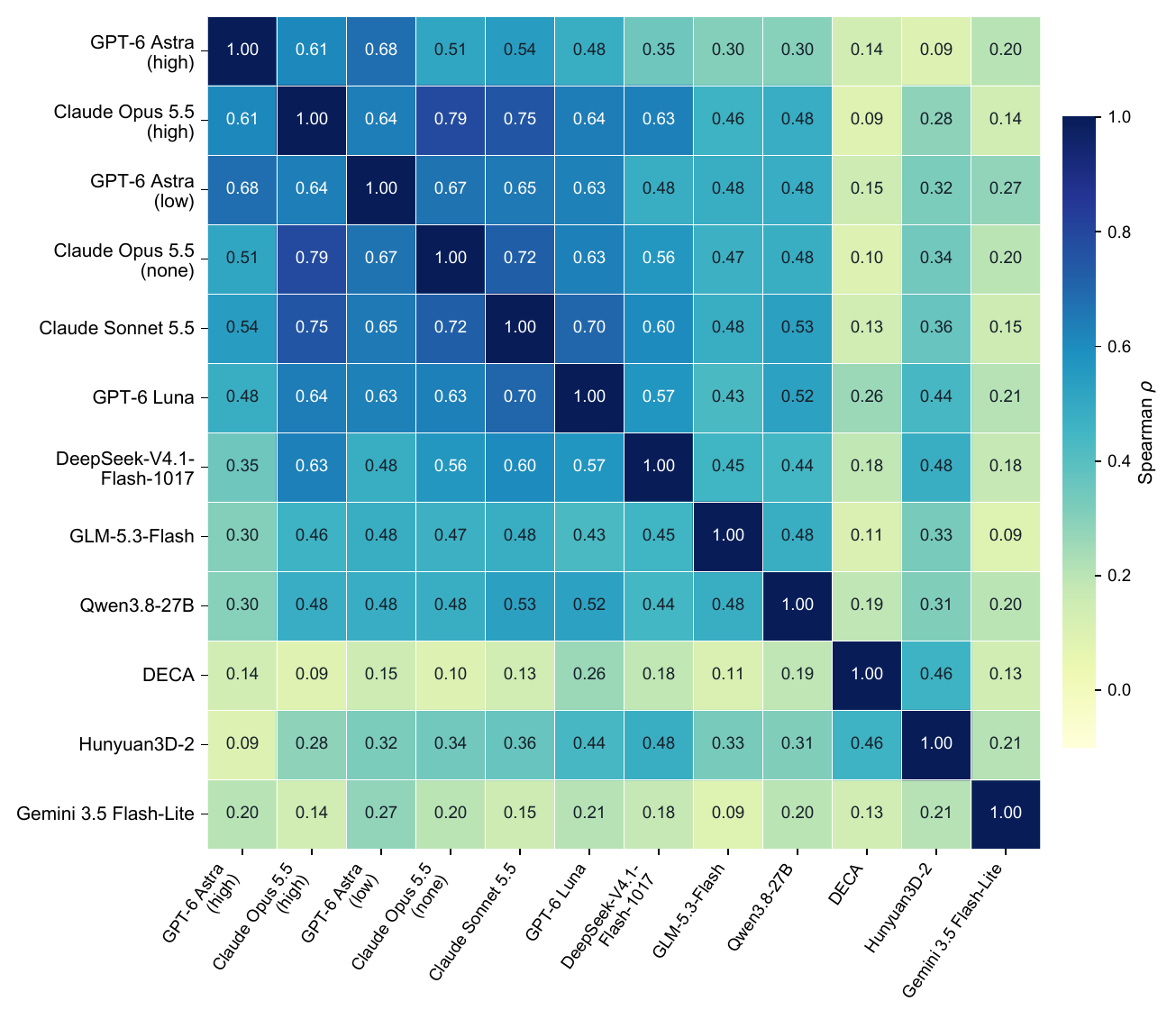}
\caption{Spearman correlation matrix of per-sample Gain. Both axes are ordered by decreasing mean Gain on the ICT100 main set. Shared targets and initializations affect correlations; these do not establish independence of difficulties across methods.}
\label{fig:s_agreement}
\end{figure}

\paragraph{Descriptive correlations.} Within each of the 10 configurations, the Spearman correlations of tokens and runtime with Gain satisfy $|\rho|\le0.21$. These observations concern naturally occurring run lengths, not fixed-budget experiments. Figure~\ref{fig:s_agreement} shows correlations of per-sample Gain.

\clearpage
\section{Diagnosis and Attempts to Improve Editing}
\label{sec:s_diagnosis}

Models with limited visual perception may misidentify which regions to edit. We test whether a stronger model's visual analysis improves reconstruction, then check whether correct visual judgments lead to correct mesh edits. Both experiments use the same 10 Easy cases drawn from ICT20.

\subsection{Testing Stronger Visual Analysis}
We replayed the visual analyses and edit plans written by GPT-6 Astra at high and low effort. In each of 40 runs, Qwen3.8-27B or DeepSeek-V4.1-Flash-1017 received the plan turn by turn, generated commands, and saw its own execution results.

Astra's analysis did not consistently improve reconstruction (Table~\ref{tab:s_plan_replay}). Qwen improved slightly with high-effort plans and declined with low-effort plans. DeepSeek declined with both. As the replay diverges from Astra's original trajectory, later analyses may refer to a different mesh. The single-step handoff controls this source of divergence.

\begin{table}[H]
\centering
\caption{Mean final Gain on the same 10 Easy cases. Own plan denotes each executor's original run. Astra's original high- and low-effort runs averaged $+0.086$ and $+0.057$, respectively.}
\label{tab:s_plan_replay}
{\linespread{1.0}\selectfont\small\renewcommand{\arraystretch}{1.14}
\begin{tabular*}{0.78\textwidth}{@{\extracolsep{\fill}}lccc@{}}
\toprule
Executor & Own plan & \makecell[cb]{Astra high\\plan} & \makecell[cb]{Astra low\\plan} \\
\midrule
Qwen & $-0.041$ & $-0.030$ & $-0.065$ \\
DeepSeek & $-0.005$ & $-0.015$ & $-0.018$ \\
\bottomrule
\end{tabular*}}
\end{table}

\subsection{Diagnosis of Geometric Editing Errors}
We replayed 20 Astra trajectories and saved the workspace before each edit to \texttt{solution.py}. Each executor received the reference image, file list, source code (up to about 60,000 characters), and Astra's current analysis and plan. It generated one batch of commands without seeing the output. All three command batches ran from the same snapshot. We measured the change in \fstrict, $\Delta F$. Of 178 handoffs, 166 entered the analysis after excluding 12 copy-only steps. We treated one unusable DeepSeek response as leaving the workspace unchanged.

Mean $\Delta F$ was positive for both Astra settings and negative for Qwen and DeepSeek. The four 95\% confidence intervals for the paired differences were below zero. The direction held after excluding steps with processes still running (Table~\ref{tab:s_step_handoff}).

\begin{figure}[H]
\centering
{\footnotesize\textcolor{cicolor}{Left to right in both panels: target, shared start, Astra, Qwen, and DeepSeek. Top: frontal view. Bottom: $90^\circ$ side view.}\par}
\vspace{5pt}
\begin{minipage}[t]{0.486\textwidth}
\centering\includegraphics[width=\linewidth,trim=0 0 0 44bp,clip]{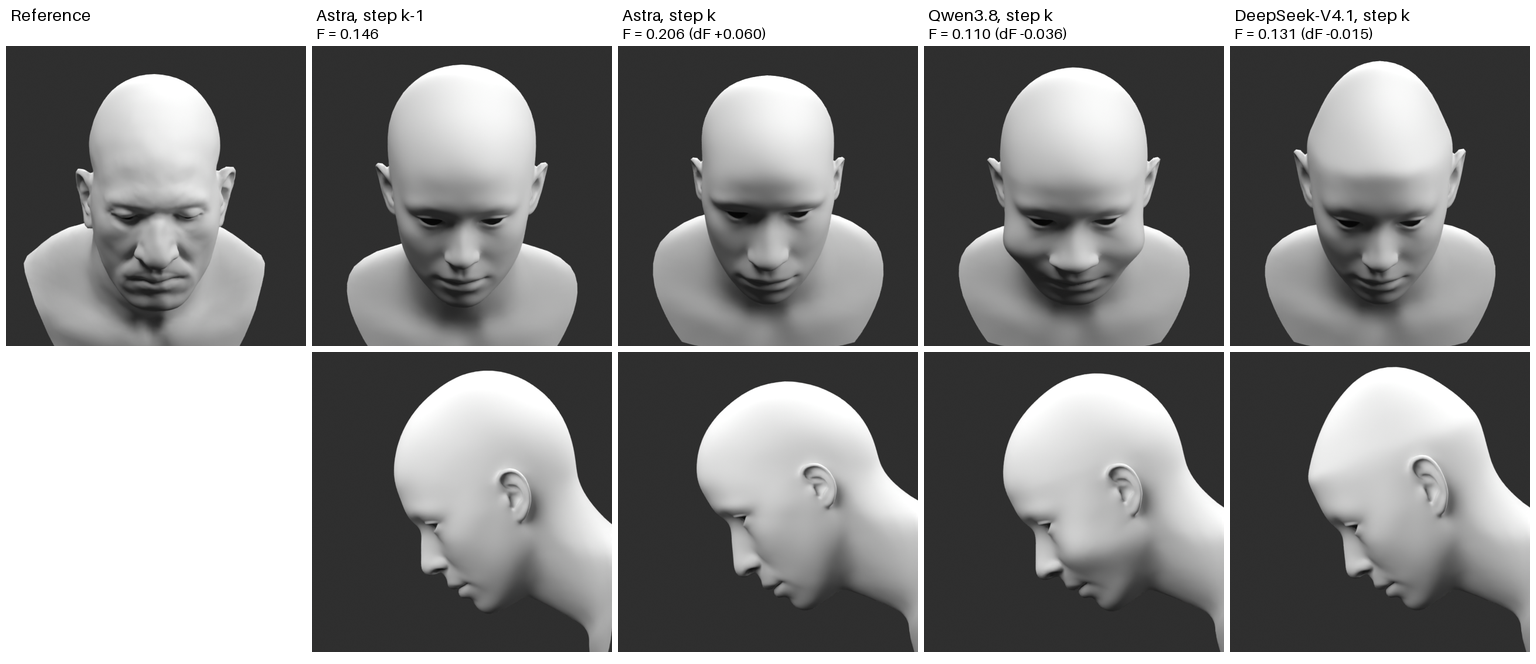}
\captionsetup{width=\linewidth}
\caption{\texttt{ict\_072}, Astra low, step 8. Shared start $F=0.146$. Astra, Qwen, and DeepSeek obtain $\Delta F$ of $+0.060$, $-0.036$, and $-0.015$.}
\label{fig:s_step_localization}
\end{minipage}\hfill
\begin{minipage}[t]{0.486\textwidth}
\centering\includegraphics[width=\linewidth,trim=0 0 0 44bp,clip]{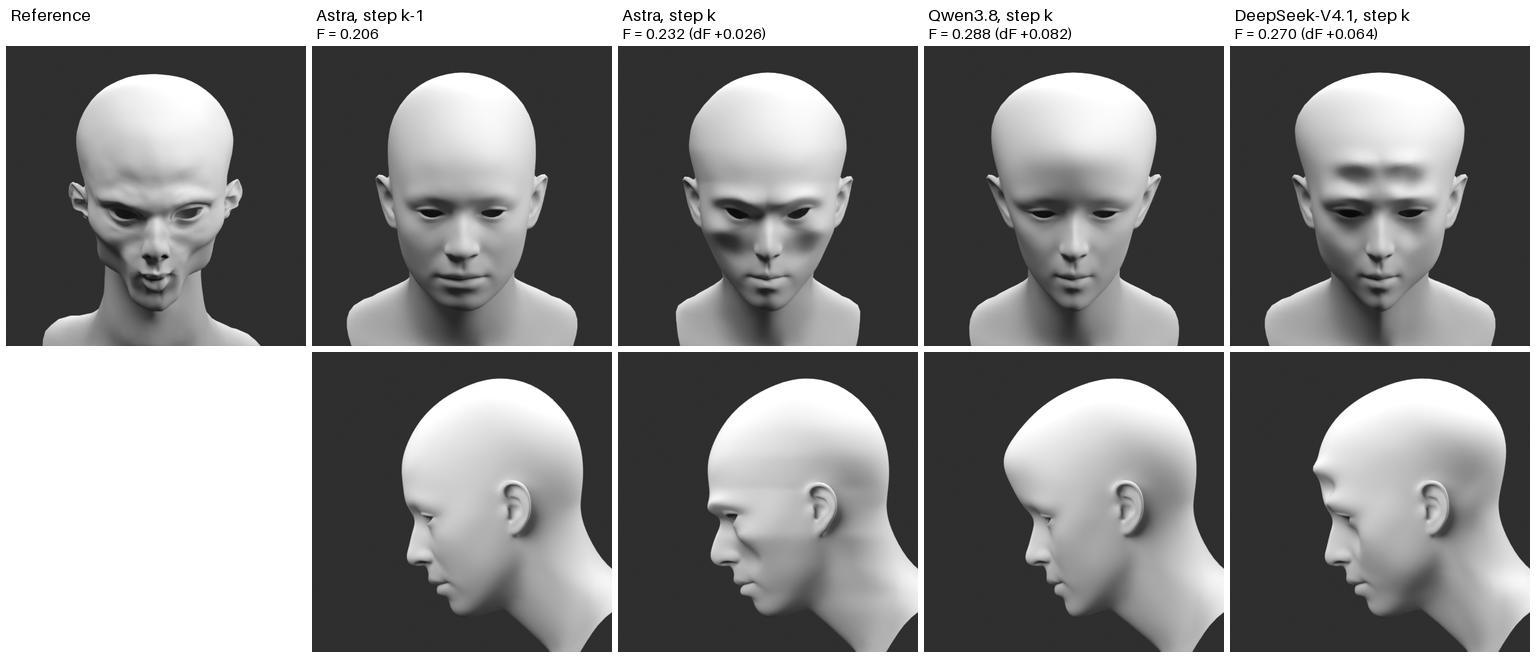}
\captionsetup{width=\linewidth}
\caption{\texttt{ict\_051}, Astra low, step 5. Shared start $F=0.206$. Qwen and DeepSeek enlarge the upper cranium more. The three $\Delta F$ values are $+0.026$, $+0.082$, and $+0.064$.}
\label{fig:s_step_positive}
\end{minipage}
\end{figure}

In Fig.~\ref{fig:s_step_localization}, Astra correctly identified that the upper cranium needed narrowing and the nose and mouth needed widening. Its code located the nose and mouth at $z\approx-0.43$ and $-0.73$. Qwen used a bounding box that included the neck and shoulders and widened the lower face and jaw. DeepSeek placed the nose and mouth edits too high. Both executors edited the wrong regions despite receiving the correct visual diagnosis.

Figure~\ref{fig:s_step_positive} shows a case where both executors did better. Astra had seen earlier measurements, while the executors relied on the handoff files for prior context. We sampled each step once and measured its immediate effect. Preparatory steps such as setting up fitting may pay off later.

\begin{table}[H]
\centering
\caption{Mean single-step $\Delta F$ and paired mean differences relative to Astra. The 95\% confidence intervals use 5,000 bootstrap resamples of target cases, keeping all steps for each selected case together.}
\label{tab:s_step_handoff}
{\linespread{1.0}\selectfont\small\setlength{\tabcolsep}{3pt}\renewcommand{\arraystretch}{1.14}
\begin{tabular*}{\textwidth}{@{\extracolsep{\fill}}lrlccc@{}}
\toprule
Astra effort & Steps & Executor & Astra $\Delta F$ & Executor $\Delta F$ & Paired difference [95\% CI] \\
\midrule
high & 130 & Qwen3.8-27B & $+0.0044$ & $-0.0063$ & $-0.0107$\ci{[-0.0182,-0.0039]} \\
high & 130 & DeepSeek-V4.1-Flash-1017 & $+0.0044$ & $-0.0044$ & $-0.0088$\ci{[-0.0137,-0.0042]} \\
low  & 36  & Qwen3.8-27B & $+0.0119$ & $-0.0118$ & $-0.0237$\ci{[-0.0450,-0.0057]} \\
low  & 36  & DeepSeek-V4.1-Flash-1017 & $+0.0119$ & $-0.0059$ & $-0.0179$\ci{[-0.0310,-0.0058]} \\
\bottomrule
\end{tabular*}}
\end{table}

\clearpage

\vspace{6pt}
\section{Qualitative Examples and Preview Trajectories}
\begin{figure}[H]
\centering
\includegraphics[width=\textwidth]{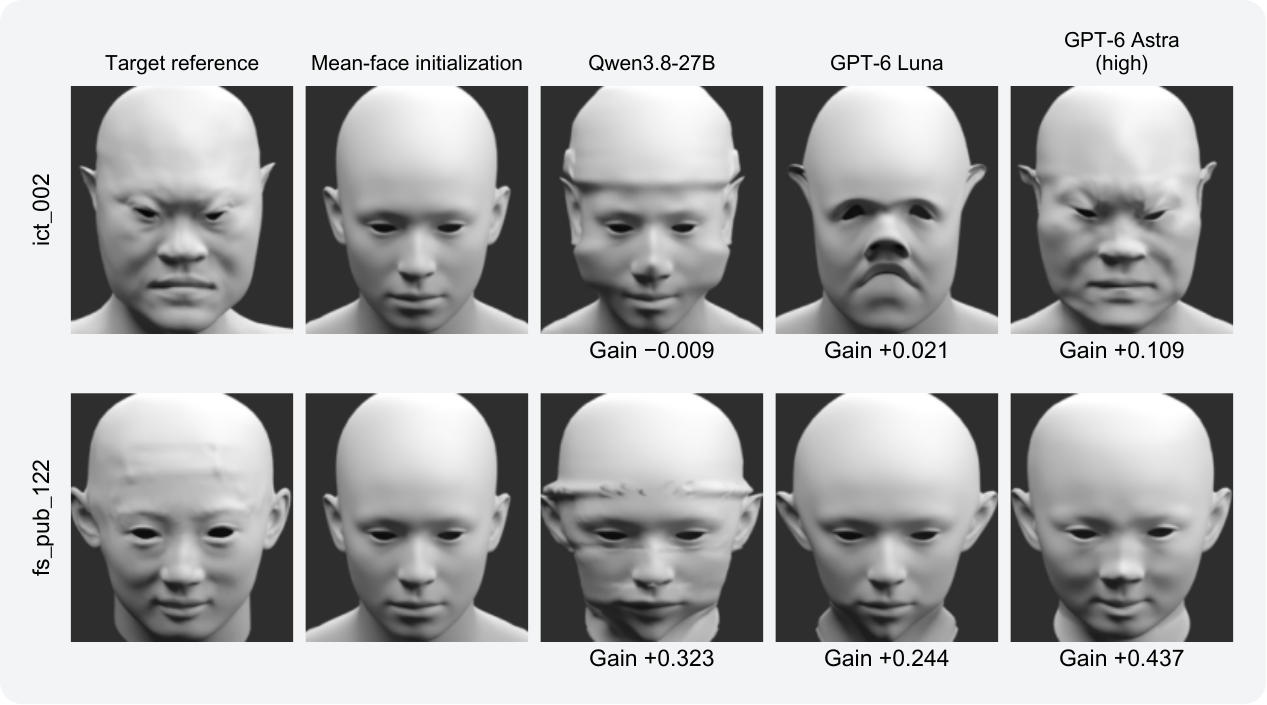}
\caption{Frontal replays of selected cases. Numbers are Gain from the existing experiments. GPT-6 Luna's mouth hole on ict\_002 and Qwen3.8-27B's transverse folds on fs\_pub\_122 can still receive positive scores; GPT-6 Astra (high) obtains higher matching scores on both cases. These examples illustrate local defects the metrics may miss and do not constitute a human perceptual agreement experiment. Complete model comparisons and preview trajectories appear in Figs.~\ref{fig:s_qual_a}--\ref{fig:s_traj}.}
\label{fig:qualitative_traj}
\end{figure}

\begin{figure}[p]
\centering
\includegraphics[width=\textwidth,height=0.86\textheight,keepaspectratio]{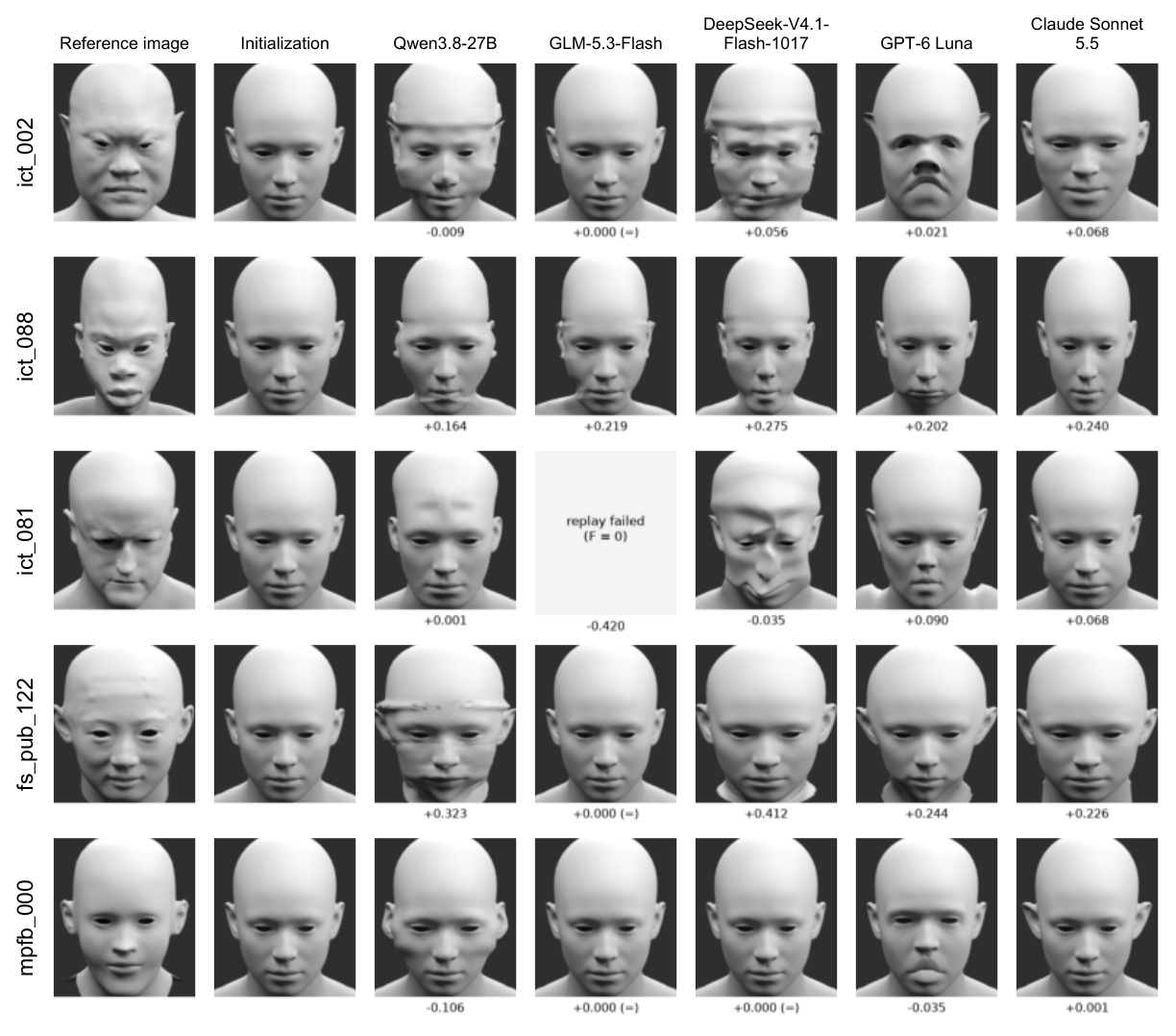}
\caption{Complete qualitative comparisons, part one: reference, initialization, and five configurations. Numbers are the original per-sample Gain values; ``(=) denotes an initialization-equivalent score (difference below $10^{-9}$), and replay failures receive zero scores.}
\label{fig:s_qual_a}
\end{figure}

\begin{figure}[p]
\centering
\includegraphics[width=\textwidth,height=0.86\textheight,keepaspectratio]{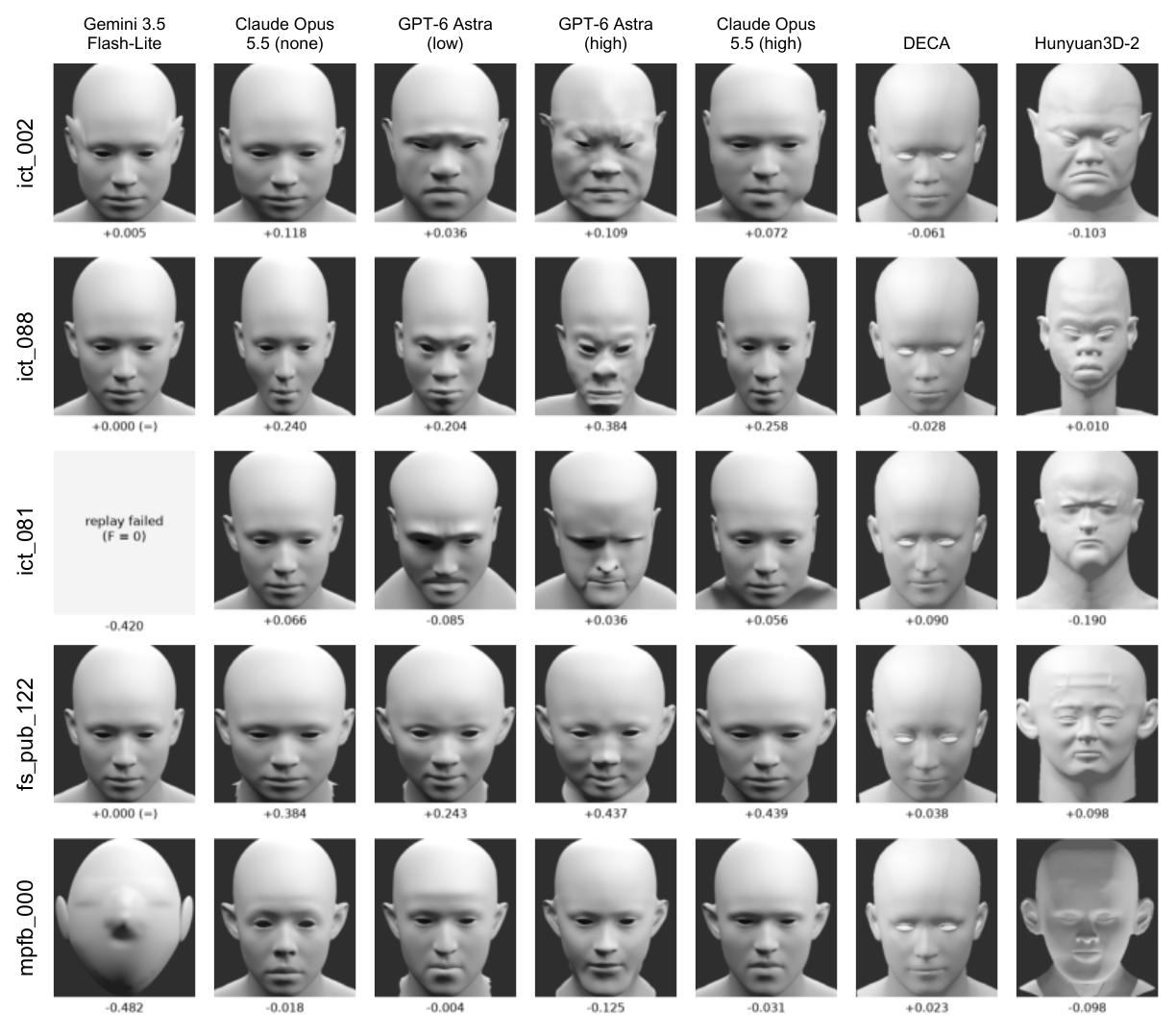}
\caption{Complete qualitative comparisons, part two: the other five agent configurations and two feed-forward references, in the same row order as the preceding figure. All images use the same frontal camera.}
\label{fig:s_qual_b}
\end{figure}

\begin{figure}[p]
\centering
\includegraphics[width=\textwidth]{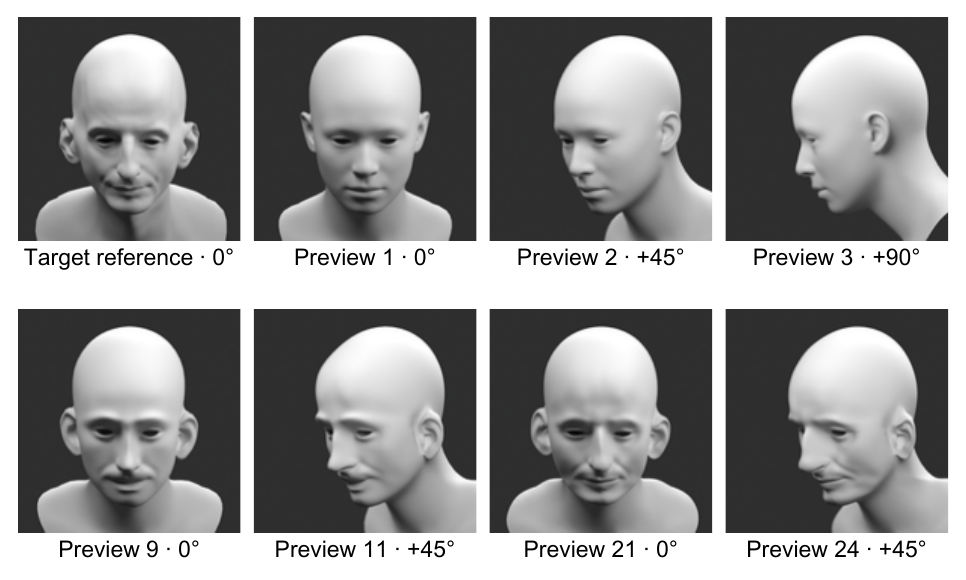}
\caption{Selected previews for GPT-6 Astra (high) on ict\_055. Indices follow the order in which images were actually opened, with the target reference first. The run took 53 minutes and 71 requests, with final Gain of +0.116. Intermediate meshes were not all scored step by step; the preview sequence is not claimed to improve monotonically.}
\label{fig:s_traj}
\end{figure}

Figure~\ref{fig:qualitative_traj} shows selected frontal replays; Figs.~\ref{fig:s_qual_a}--\ref{fig:s_traj} provide complete model comparisons and a multiview preview example. The preview sequence here shows only that the model requested different viewpoints; the main text additionally reports post-hoc replay scores of saved scripts from selected long-horizon trajectories. We did not score every intermediate version across all runs, and the human comparison covers only a fixed candidate set.

\clearpage
\section{Protocol Details and Reproducibility Scope}
The reference camera is orthographic, with resolution $512\times512$ and scale 2.6. It is located at $6(\sin\psi\cos\theta,-\cos\psi\cos\theta,\sin\theta)$ and looks at the origin; the input image uses $\psi=\theta=0$. Only the region $z\ge-1.05$ is scored. Final scripts are replayed in a fresh Blender 5.2.1 process with a 120-second limit and at most 500,000 triangles. Renderable MESH/CURVE/SURFACE/FONT/META objects are converted to triangle meshes for evaluation; temporary files are not delivered to the replay environment.

The evaluator samples 100,000 points in proportion to area from each clipped prediction and ground-truth surface, approximates surface distances using bidirectional nearest sampled points, and sets thresholds relative to the clipped ground-truth bounding-box diagonal. Sampling density may affect this discrete approximation at strict thresholds; deterministic sampling ensures repeatable scores. The ground-truth script Oracle self-check of 1.000 validates the pipeline and sufficiently validates numerical accuracy across different topologies.

After regression, the DECA baseline uses a neutral FLAME shape with expression and pose set to zero and canonicalizes it using five landmarks on the prediction itself; this choice may limit performance on expressive targets. Hunyuan3D-2 uses a fixed rotation, input frontal-silhouette placement, and a mean-head depth prior.

\vspace{10pt}
\begin{figure}[H]
\centering
\includegraphics[width=\textwidth]{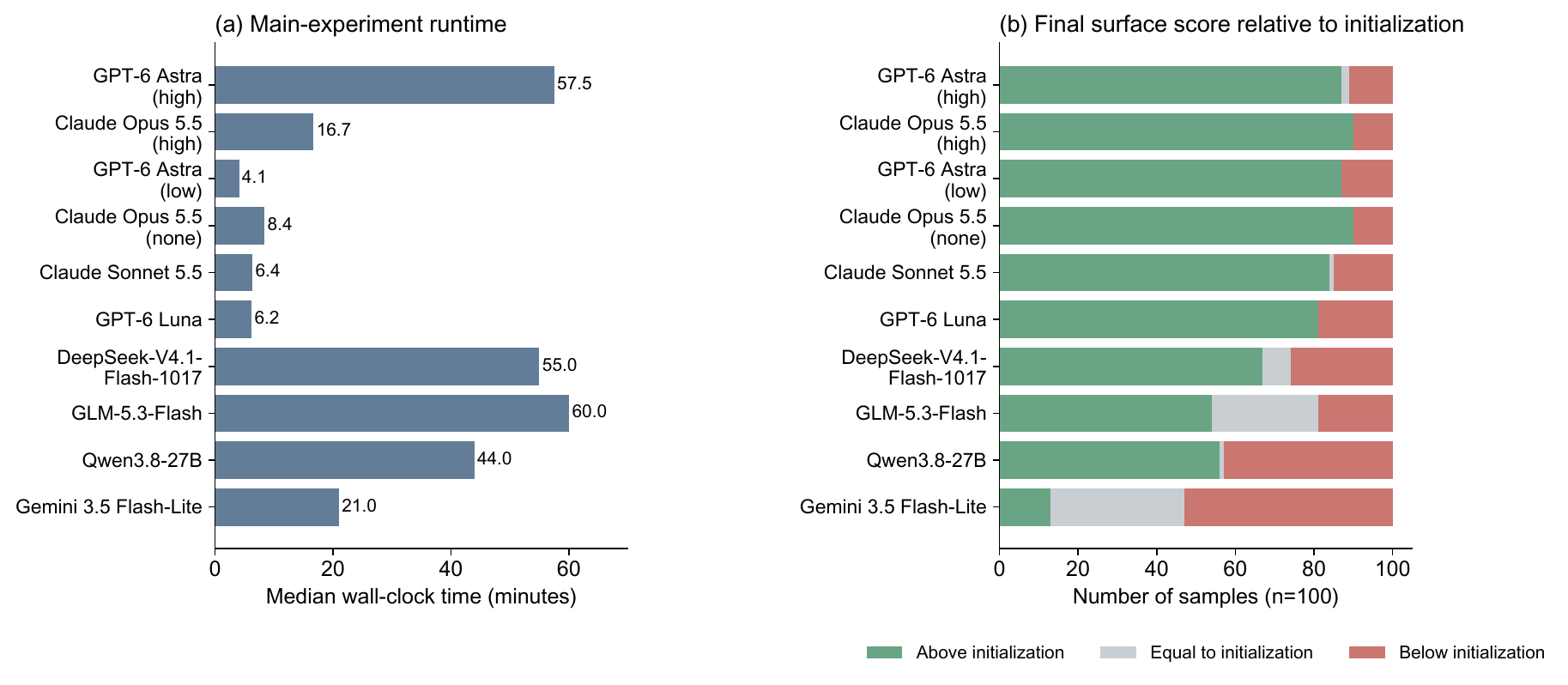}
\caption{Interaction duration and final-score states. A score is equal to initialization when its F-score differs by less than $10^{-9}$; this does not establish byte-identical scripts. Complete cost, token, and correlation statistics appear in Table~\ref{tab:s_effort}.}
\label{fig:s_effort}
\end{figure}

\clearpage

% Additional appendix material; existing sections and numbering remain unchanged.
\clearpage
\section{Test Prompts}
\label{sec:s_prompts}
Figure~\ref{fig:s_prompts} presents the English task prompts actually issued, using the ICT single-frontal-view task as an example and reformatting the original text as bullets. Default uses the shared instruction; Easy appends the initialization instruction. The prompts come from \texttt{INSTRUCTION} and \texttt{EASY\_ANSWER} in the data-construction script, with camera and budget placeholders expanded according to the experimental settings.

\vspace{4pt}
\begin{figure}[H]
\centering
\begingroup
\ifdefined\nolinenumbers\nolinenumbers\fi
\small\linespread{1.08}\selectfont
\begin{promptbox}{Shared task prompt --- Default and Easy}
{\normalsize\bfseries\textcolor{prompttitle}{Reconstruct a head in Blender}}\par\vspace{3pt}
Reconstruct the head surface shown in the reference images in Blender. Match its 3D geometry in the specified canonical frame. Include face, cranium and ears; exclude hair, eyebrows, eyeballs, teeth, tongue, eyelashes, clothing and accessories.

\vspace{5pt}\noindent{\bfseries\textcolor{prompttitle}{Inputs}}\par\vspace{1pt}
\begin{itemize}[leftmargin=1.35em,itemsep=2pt,topsep=2pt]
\item \texttt{/app/reference/view\_000.png}: yaw $0^\circ$, pitch $0^\circ$.
\item Every reference is a 512x512 orthographic render, ortho\_scale 2.6, looking at the origin. A camera with yaw $\psi$ and pitch $\theta$ sits at $6\cdot(\sin\psi\cos\theta, -\cos\psi\cos\theta, \sin\theta)$; image up is $+Z$ at pitch 0. The same data is in \texttt{/app/task.json}.
\item Canonical frame: $+X$ is subject right, $+Z$ up, $-Y$ forward; crown-to-chin height is 2, centered at the origin.
\item No ground truth and no other views of the target exist anywhere in this environment.
\end{itemize}

\vspace{5pt}\noindent{\bfseries\textcolor{prompttitle}{Environment}}\par\vspace{1pt}
\begin{itemize}[leftmargin=1.35em,itemsep=2pt,topsep=2pt]
\item Linux, no network, no GPU. You may write anywhere under \texttt{/app}, \texttt{/tmp} and your home directory.
\item \texttt{blender} (5.2.1, background mode only; Cycles on CPU). Its bundled Python has numpy.
\item \texttt{bench-preview SCRIPT [--view YAW,PITCH ...]} runs SCRIPT from an empty factory scene exactly as the grader does and renders grey clay images of the result (the reference views by default). It prints JSON with success, triangle count and image paths.
\item \texttt{bench-check} does the same for \texttt{/app/submission/solution.py}.
\end{itemize}

\vspace{5pt}\noindent{\bfseries\textcolor{prompttitle}{Answer}}\par\vspace{1pt}
\begin{itemize}[leftmargin=1.35em,itemsep=2pt,topsep=2pt]
\item Write \texttt{/app/submission/solution.py}: one complete, self-contained Blender Python script that builds the head from an empty factory scene. It must not read any other file; the grader runs it alone in a fresh Blender.
\item The grader keeps only render-visible MESH, CURVE, SURFACE, FONT and META geometry and allows at most 500,000 triangles and 120 seconds.
\item Whatever that file contains when your session ends is your answer. There is no confirmation step. A missing or failing script scores 0.
\item You have 60 minutes in total.
\end{itemize}
\end{promptbox}
\vspace{6pt}
\begin{promptbox}{Additional prompt --- Easy (ICT tasks)}
{\bfseries\textcolor{prompttitle}{Starting point}}\par\vspace{1pt}
\begin{itemize}[leftmargin=1.35em,itemsep=2pt,topsep=2pt]
\item \texttt{/app/submission/solution.py} already exists and builds the \textbf{average ICT head} in this canonical frame --- the mean of the other 99 ICT identities, so it is a plausible head but not this subject. It is a normal, complete answer: run it with \texttt{bench-check} to see it.
\item Your job is to edit it toward the subject in the reference image and leave the result in that same file. Starting from it instead of an empty scene is deliberate: the task is how far you can move this head toward the target, not how well you can build one from nothing.
\item \texttt{/logs/verifier/reward.json} also reports \texttt{init\_f005} / \texttt{init\_f02}, the score of this starting mesh, so a run can be reported as a gain over it.
\end{itemize}
\end{promptbox}
\endgroup
\caption{Task prompts used in the benchmark: the shared instruction and the additional starting-point instruction for Easy ICT tasks.}
\label{fig:s_prompts}
\end{figure}

\clearpage
\section{Frontal and Side Previews of All Samples}
\label{sec:s_all_previews}
\begin{figure}[H]
\centering
\includegraphics[page=1,width=\textwidth,height=0.81\textheight,keepaspectratio]{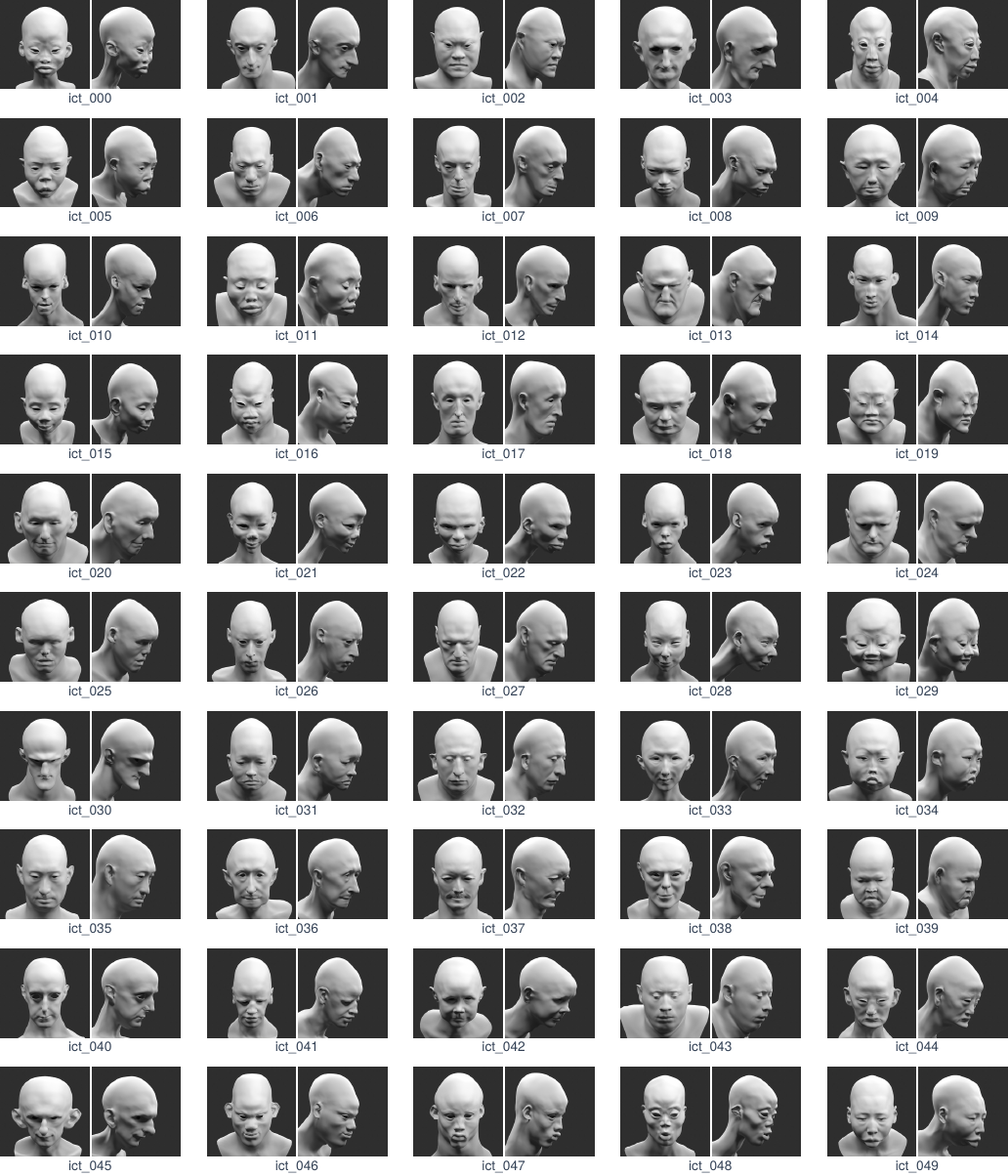}
\caption{Paired frontal (left) and side (right, yaw $-60^\circ$) previews of ICT cases 000--049. Case IDs increase from left to right and top to bottom.}
\label{fig:s_all_ict}
\end{figure}

\begin{figure}[p]
\centering
\includegraphics[page=2,width=\textwidth,height=0.88\textheight,keepaspectratio]{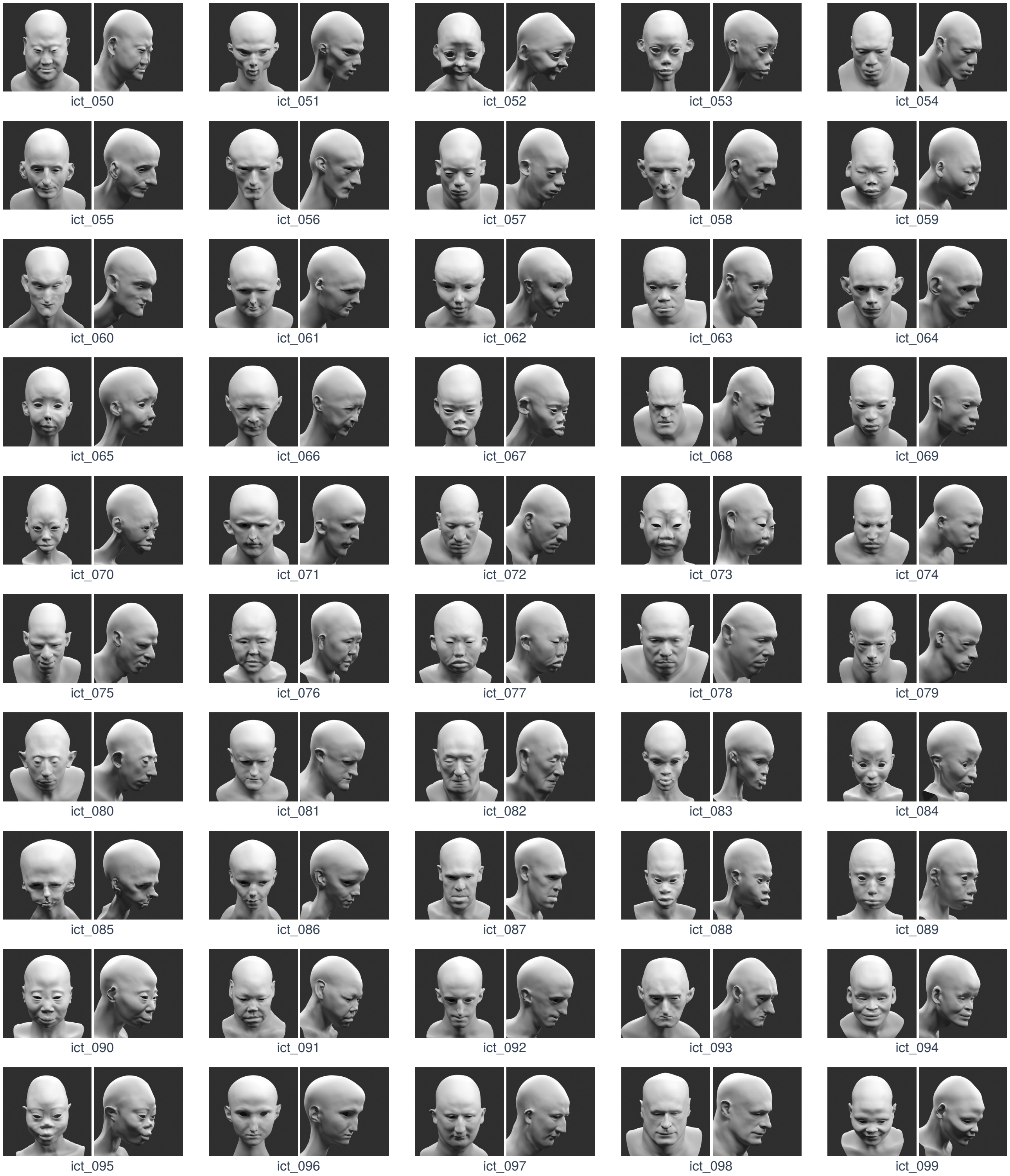}
\caption{Paired frontal and side previews of ICT cases 050--099, using the same view convention as the preceding figure.}
\label{fig:s_all_ict_b}
\end{figure}

\begin{figure}[H]
\centering
\includegraphics[width=\textwidth]{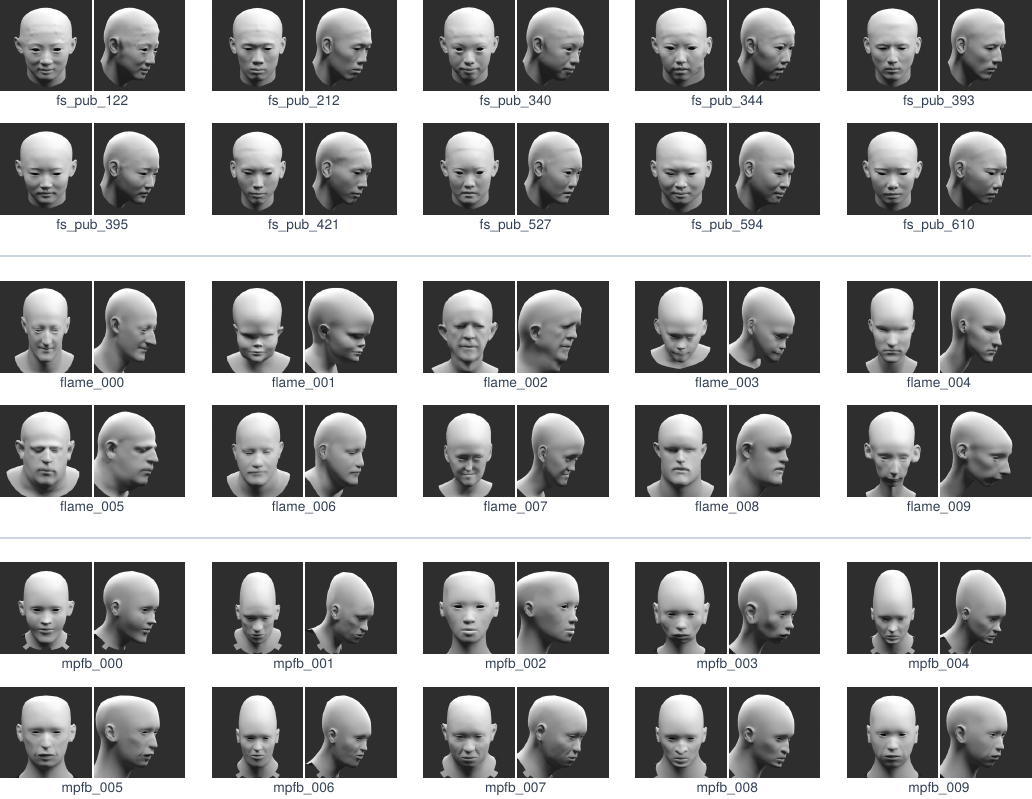}
\caption{Paired frontal and side previews of all 30 cross-domain cases. Each source occupies two consecutive rows: FaceScape, FLAME, then MPFB.}
\label{fig:s_all_crossdomain}
\end{figure}

Figures~\ref{fig:s_all_ict}--\ref{fig:s_all_crossdomain} cover all 130 cases: 100 ICT, 10 FaceScape, 10 FLAME, and 10 MPFB. For each case, the left image is frontal (yaw $0^\circ$) and the right image is a side view (yaw $-60^\circ$), both at pitch $0^\circ$. Side views are held-out views shown only to illustrate the dataset; agents receive only the frontal target reference during testing.

\clearpage

\clearpage
\subsection{Cross-Domain Samples: FaceScape, FLAME, and MPFB}
\vspace{4pt}

\end{document}